\documentclass{article} %
\usepackage{iclr2027_conference,times}

\usepackage{hyperref}
\usepackage{url}
\usepackage[utf8]{inputenc} %
\usepackage[T1]{fontenc}    %
\usepackage{hyperref}       %
\usepackage{url}            %
\usepackage{booktabs}       %
\usepackage{amsfonts}       %
\usepackage{nicefrac}       %
\usepackage{microtype}      %
\usepackage{xcolor}         %
\usepackage{natbib}
\usepackage{amsmath}
\usepackage{amssymb}
\usepackage{mathtools}
\usepackage{amsthm}
\usepackage{multirow}
\usepackage{placeins}
\providecommand{\msd}[2]{\ensuremath{#1{\scriptstyle\,\pm\,#2}}}
\providecommand{\vsd}[2]{\begin{tabular}[t]{@{}r@{}}#1\\[-2pt]{\scriptsize($\pm$\,#2)}\end{tabular}}

\providecommand{\papertablesetup}{%
  \setlength{\aboverulesep}{0pt}%
  \setlength{\belowrulesep}{0pt}%
  \renewcommand{\arraystretch}{1.1}%
}

\usepackage{subcaption}
\usepackage{tablefootnote}
\usepackage{enumitem}
\usepackage{wrapfig}
\usepackage{needspace}
\usepackage{makecell}
\usepackage{kotex}
\usepackage[linesnumbered,ruled]{algorithm2e}

\SetCommentSty{mycommfont}
\usepackage{subfiles}
\usepackage{titletoc}
\newcommand\DoToC{%
  \startcontents
  \printcontents{}{1}{\vskip3pt\hrule\vskip5pt}
}
\usepackage{colortbl}
\definecolor{gainsboro}{RGB}{233,233,233}

\usepackage{longtable}
\usepackage{booktabs}

\usepackage{xcolor}
\usepackage{pifont}
\usepackage{graphicx}

\usepackage[most]{tcolorbox}
\usepackage{booktabs}
\usepackage{siunitx}
\definecolor{additiongreen}{RGB}{0,120,65}
\newcommand{\proposed}{\textsf{CondMSP}}
\title{CondMSP: Condition-Informed and Resource-Aware Inorganic Materials Synthesis Planning}

\author{Heewoong Noh$^{1}$, Gyoung S. Na$^{1}$, Elton Pan$^{2}$, Namkyeong Lee$^{1}$, Chanyoung Park$^{1}$\thanks{Corresponding author (\texttt{cy.park@kaist.ac.kr})} \\
\\
$^1$KAIST, $^2$MIT
}

\iclrfinalcopy %
\begin{document}

\maketitle
\lhead{Preprint}

\begin{abstract}

Materials synthesis planning (MSP) connects AI-driven materials discovery to
experimental realization by proposing precursors, synthesis operations, and
processing conditions for a target material. Existing approaches predict
these components separately or in pairs and treat conditions only as a
prediction target. Consequently, MSP remains centered on recipe prediction: it leaves
unexplored how conditions can serve as process context for constructing
synthesis routes, and it does not address which candidate routes should be
executed under limited experimental resources. We propose \proposed{}, which
extends MSP from recipe prediction to execution-aware planning. \proposed{}
predicts temperature, time, and atmosphere by combining a joint predictor
with variable-specific experts, and then uses these predictions both as
process context, grounding the generation and retrieval of operation
sequences, and as execution requirements, specifying which thermal stages
can share a furnace run and at what cost. This specification is held
fixed across all compared methods; within it, \proposed{} ranks routes across
precursor sets by a common utility and, under furnace budgets, jointly selects
routes and allocates their thermal stages by constrained optimization.
\proposed{} improves condition and operation prediction and, under this
shared constraint, raises reference-recipe recovery from 27.9\% to 32.9\% at
three simulated furnace runs per target and from 34.7\% to 40.0\% at 30
trials, while a new evaluation set of 2,344 targets supports generalization
beyond oxide-dominated training chemistry.\looseness-1
\end{abstract}

\begingroup
\setlength{\textfloatsep}{7pt plus 1pt minus 1pt}
\setlength{\floatsep}{5pt plus 1pt minus 1pt}
\setlength{\intextsep}{5pt plus 1pt minus 1pt}
\renewcommand{\topfraction}{0.95}
\renewcommand{\bottomfraction}{0.9}
\renewcommand{\textfraction}{0.05}
\renewcommand{\floatpagefraction}{0.85}
\setcounter{topnumber}{4}
\setcounter{bottomnumber}{3}
\setcounter{totalnumber}{6}
\captionsetup{font=small,skip=3pt}
\setlist[itemize]{topsep=2pt,itemsep=0pt,parsep=0pt,partopsep=0pt}
\abovedisplayskip=5pt plus 1pt minus 1pt
\belowdisplayskip=5pt plus 1pt minus 1pt
\abovedisplayshortskip=0pt plus 1pt
\belowdisplayshortskip=3pt plus 1pt minus 1pt

\renewcommand{\vsd}[2]{\begingroup\renewcommand{\arraystretch}{0.9}%
\begin{tabular}[t]{@{}r@{}}#1\\[-2pt]{\scriptsize($\pm$\,#2)}\end{tabular}\endgroup}

\setlength{\parskip}{3pt}

\section{Introduction}
\label{sec: Introduction}
Materials synthesis planning (MSP) translates AI-driven materials discovery into experimental realization by recommending a synthesis procedure: (1) the precursors, (2) operation sequences, and  (3) processing conditions needed to synthesize a target material. For inorganic materials, these choices are interdependent: the same target can be synthesized from different precursor combinations, while thermal and atmospheric conditions influence the phases that eventually form\citep{rao1997new, west2022solid}. For example, even with the same precursors, an unsuitable heating temperature can leave a reaction incomplete or favor unwanted phases \citep{huo2022machine}. Consequently, choosing one component of the synthesis plan constrains the others, so that none can be determined well in isolation. Moreover, limited experimental trial budgets and equipment availability require choosing which of the many possible routes to test. Owing to this complexity, generalized synthesis rules remain difficult to establish, and route design has long relied on empirical knowledge and expert judgment\citep{jansen2002concept}.

Existing AI-based methods have largely addressed these components individually: precursor prediction (PP) proposes precursor sets for a target \citep{he2023precursor,noh2024retrieval,pan2026synreason,prein2025language,prein2025retro,prein2026benchmarking}; synthesis operation prediction (SOP) generates operation sequences from the target and its precursors \citep{na2023artificial}; and synthesis condition prediction (SCP) predicts conditions such as heating temperature and time \citep{huo2022machine,karpovich2023interpretable,prein2024reaction,pan2026diffsyn}. Predicting components in isolation, however, does not account for how they depend on one another, and recent methods therefore connect several of them: SPENDE predicts operation sequences and their associated engineering conditions for thermoelectric materials \citep{na2023artificial}, while MSP-LLM pairs predicted precursor sets with operation sequences to construct candidate synthesis routes \citep{noh2026msp}. These approaches improve component prediction, but leave two questions open: (1) they either produce conditions as outputs or do not predict them at all, leaving unexplored \textit{how conditions can serve as process context for SOP}, even though conditions and operations are interdependent; and (2) they stop at proposing candidates for a target, leaving unaddressed \textit{how a fixed experimental budget should be allocated across the resulting synthesis routes.}

First, synthesis conditions can provide process context for operation prediction. Operation sequences specify what is done to the precursors, while conditions such as temperature, time, and atmosphere describe the regime in which those operations are carried out. In practice, the two are closely linked; high-temperature treatments, for example, accompany operations such as sintering. Analyses of text-mined recipes have likewise reported systematic relationships between heating procedures and their temperatures \citep{huo2022machine}. Making the anticipated processing regime explicit therefore gives operation prediction a physically meaningful intermediate to build on, linking predicted operations to concrete process parameters.

Second, existing MSP stops at proposing candidates and thus remains primarily a recipe-prediction problem: it asks which synthesis routes are plausible, not which of them should be executed under limited experimental resources. For a single target, PP can propose several precursor sets, each with multiple candidate operation sequences, yielding a pool of alternative routes, and allocating a finite budget across them is a planning decision distinct from predicting their components: it requires weighing how strongly each route is supported by the predictions against the resources required for its execution. Batch-processing platforms such as self-driving laboratories make the resource side explicit, since equipment use depends on which experiments are executed together \citep{szymanski2023autonomous}. Which routes can share furnace capacity depends on their synthesis conditions: routes with compatible temperature, time, and atmosphere requirements can be batched together, whereas repeated heating introduces additional thermal stages that each require capacity. The resource cost of a route therefore depends on which other routes are selected with it. With a trial budget alone, selection reduces to ranking routes by a utility that must be comparable across precursor sets; once routes share furnace capacity, it becomes a budget allocation problem that ranking alone does not solve. Formulating it as such for each target extends MSP from predicting recipes to planning their execution.\looseness=-1

In this work, we propose \proposed{}, an MSP framework that moves from predicting recipes to planning their execution: rather than treating synthesis conditions only as prediction targets, it predicts them and uses them both as process context for constructing synthesis routes and as execution requirements for deciding which routes to attempt under limited experimental resources. \proposed{} first predicts temperature, time, and atmosphere from the target--precursor pair by combining a joint condition LLM with variable-specific experts. The predicted conditions then enter SOP through two complementary paths: condition-grounded generation uses them as context for proposing operation sequences, while condition-aware retrieval weights precursor-overlap evidence from recorded syntheses by condition similarity, and their fusion yields a unified SOP ranking. Finally, \proposed{} combines PP confidence and SOP rank into a route utility and selects routes by constrained optimization under trial, furnace-run, and furnace-hour budgets, with predicted conditions and thermal stages determining execution costs. Extensive experiments show that \proposed{} improves reference-recipe recovery (exact match of precursor set and operation sequence) from 27.9 to 32.9\% under three simulated furnace runs per target and from 34.7 to 40.0\% at 30 trials, compared with the strongest baseline. These gains have two sources: as process context, predicted conditions raise operation-sequence coverage; route selection spreads trials across precursor sets by a common utility and, under furnace budgets whose compatibility and cost the predicted conditions define for all methods, recovers routes that utility ranking alone misses. On a new dataset of 2,344 targets, this advantage persists without retraining, supporting generalization beyond the oxide-dominated training distribution.\looseness=-1

Our contributions are as follows:
\begin{itemize}[leftmargin=10pt]
    \item We formulate MSP as an execution-aware planning problem that extends beyond recipe prediction to selecting which synthesis routes to execute under limited experimental resources. Predicted synthesis conditions serve both as process context for route construction and as execution requirements that define the resource constraint under which routes are selected.
    \vspace{0.6em}
    \item We develop condition-aware SOP that uses predicted conditions to guide operation-sequence generation and to weight evidence from recorded syntheses, fused into a unified ranking, and a resource-aware selector that allocates trials across routes, balancing precursor sets against SOP candidates within each set, under condition-dependent furnace constraints.
    \vspace{0.6em}
    \item Extensive experiments demonstrate that \proposed{} improves synthesis condition and operation prediction over existing methods and achieves higher reference-recipe recovery under limited trial and simulated furnace budgets. We further construct a new evaluation dataset of 2,344 targets to assess generalization under a shifted chemistry distribution.
\end{itemize}
\vspace{-0.3em}

\vspace{-0.8em}
\section{Preliminaries}
\vspace{-0.4em}
\label{sec: Preliminaries}
\subsection{Materials Synthesis Planning (MSP)}
\vspace{-0.4em}

Materials synthesis planning (MSP) aims to identify how a target material can be experimentally synthesized. Given a target material $m$, a candidate synthesis plan comprises a precursor set $p$, a synthesis operation sequence $o=[o_1,\ldots,o_L]$, and the corresponding synthesis conditions $c$. Following prior work \citep{he2023precursor,noh2024retrieval, prein2025language}, a target is specified by its composition alone, without structural information. In this work, a synthesis procedure can be factorized into two parts: a synthesis route $r=(p,o)$, and its process conditions $c$. An experimental plan specifies which candidate routes to attempt and assigns their heating steps to furnace runs (Section \ref{sec:resource_planning}). A thermal stage is a heating step within a route, indexed in execution order; a furnace run processes a batch of at most $C$ samples with compatible stage requirements, where $C$ denotes furnace capacity. Our framework produces the experimental plan in a single shot, without feedback from experimental outcomes.
\subsection{Precursor and Synthesis Operation Prediction (PP \& SOP)}
\vspace{-0.4em}
We build upon MSP-LLM \citep{noh2026msp}, a recent LLM-based framework that organizes materials synthesis planning into precursor prediction (PP) and synthesis operation prediction (SOP). PP identifies candidate combinations of starting materials for a target material $m$, producing a ranked list of precursor sets $\mathcal{P}_m=(p_1,\ldots,p_K)$. The autoregressive PP model generates each precursor set as a token sequence, so distinct decoded sequences can represent the same unordered set. For each precursor candidate $p_k$, SOP predicts how the precursors should be processed by conditioning on both the target and the precursor set, yielding a ranked list of operation sequences $\mathcal{O}_{m,k}=(o_{k1},\ldots,o_{kJ})$. Operations are drawn from seven labels adopted from the same framework: \emph{mixing}, \emph{heating}, \emph{sintering}, \emph{shaping}, \emph{drying}, \emph{quenching}, and \emph{annealing}, where sintering and annealing are labeled separately from other heating treatments. Pairing $p_k$ with an operation sequence $o_{kj}$ defines a candidate synthesis route $r_{kj}=(p_k,o_{kj})$. We adopt the framework's PP component and extend SOP with synthesis-condition information.\looseness-1

\subsection{Synthesis Condition Prediction (SCP)}
\vspace{-0.4em}
The synthesis conditions associated with a route are denoted by $c=(T,t,a)$, where $T$ denotes the maximum heating temperature, $t$ denotes the maximum heating time, and $a$ denotes the synthesis atmosphere (i.e., air/nonair). Because stage-level annotations are incomplete, conditions are summarized at the record level: these maxima provide a common basis for comparing thermal requirements across procedures with different numbers of heating stages. Conditions are not given when a route is proposed, and for a target--precursor pair $(m,p_k)$ they are only partially annotated in the training data. We therefore predict them as $\hat{c}_k=f_{\mathrm{cond}}(m,p_k)$, which does not require operation annotations and can thus be trained on records unusable for SOP (Appendix~\ref{app:dataset}). The predicted conditions then serve as process context for operation prediction and, in resource-aware planning, as the execution requirements that define furnace compatibility and cost for every compared method.\looseness-1

\vspace{-0.7em}
\section{Proposed Method: \proposed}
\vspace{-0.4em}
\label{sec: Methodology}
\begingroup

\proposed{} (Figure \ref{fig:method_overview} (a)) performs synthesis planning in three stages. First, for each precursor candidate \(p_k\) of a target material \(m\), it predicts synthesis conditions \(\hat{c}_k\) using a hybrid condition model that combines shared and variable-specific predictions (Figure \ref{fig:method_overview} (b); Section \ref{Sec: synthesis condition modeling}). 
Second, it uses these predicted conditions to guide both generation and retrieval of synthesis operation sequences, whose candidates are fused into a unified SOP ranking (Figure \ref{fig:method_overview} (c); Section \ref{Sec: Condition-Integrated Synthesis Operation Prediction}). 
Third, \proposed{} constructs synthesis routes from precursor–operation pairs and allocates the available experimental budget across these routes, accounting for route utility and shared furnace constraints (Figure \ref{fig:method_overview} (d); Section \ref{sec:resource_planning}). 
The first two stages are applied independently to each precursor candidate, while the final planning stage jointly selects routes across precursor branches.

\begin{figure}[!t]
\centering
\includegraphics[width=0.85\linewidth]{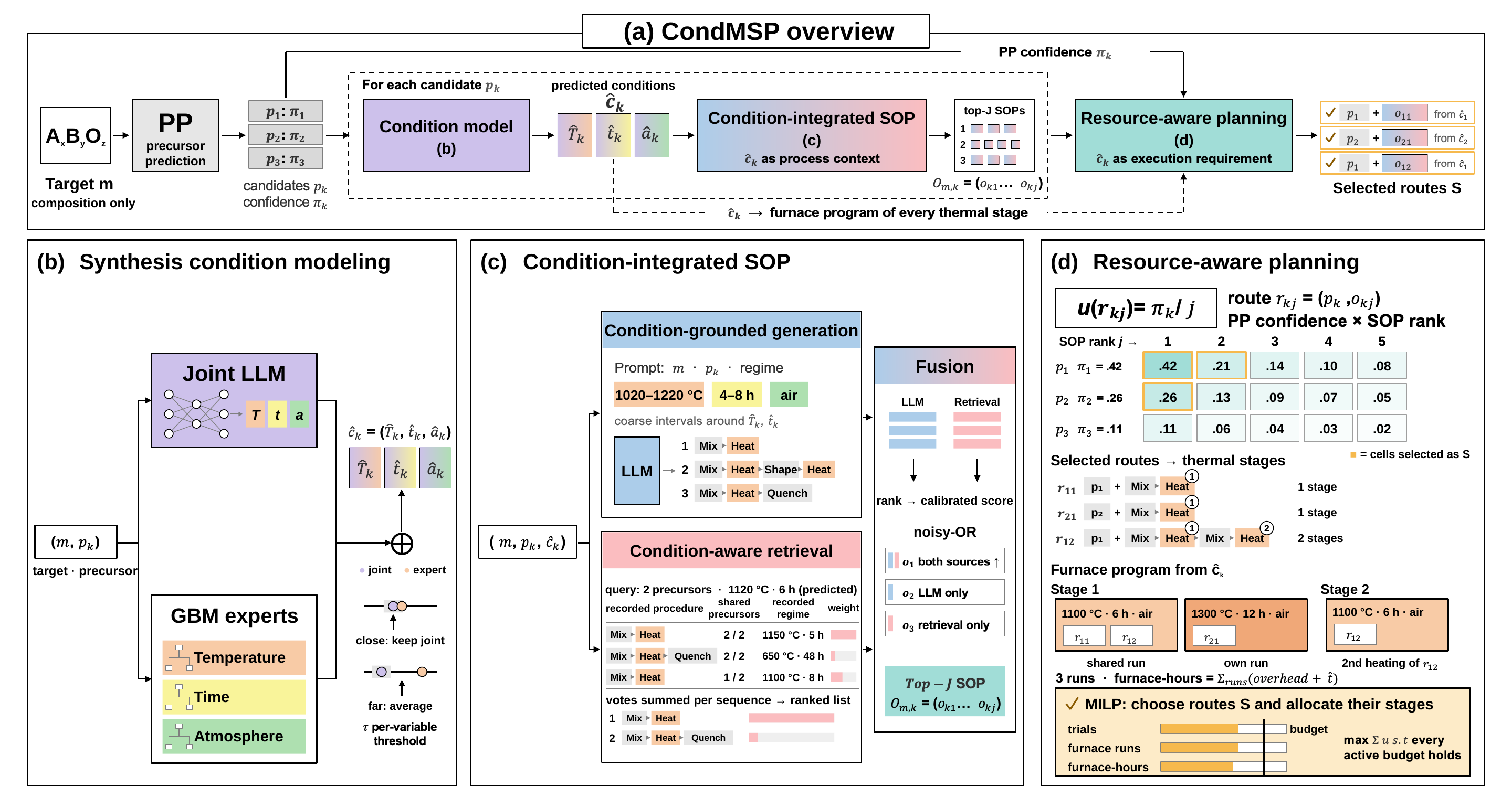}
\caption{{{Overview of \proposed{}.} (a) Predicted conditions connect operation prediction with resource-aware planning. (b) A joint LLM and variable-specific experts predict synthesis conditions. (c) Condition-grounded generation and condition-aware retrieval are fused into an SOP ranking. (d) MILP selects routes and allocates thermal stages under resource budgets. Recipes and numerical values are illustrative.}}
\label{fig:method_overview}
\end{figure}

\subsection{Synthesis Condition Modeling}
\label{Sec: synthesis condition modeling}

Conditions describe the regime in which operations are carried out, and \proposed{} uses them both to ground operation prediction and to characterize execution requirements. Temperature, time, and atmosphere are process variables of the same synthesis procedure, which supports a shared representation for their prediction. At the same time, they differ in character: temperature and time are continuous, but time spans a wide range, so the same absolute error is more consequential at short durations than at long ones, and atmosphere is categorical and imbalanced. 
We therefore combine a joint predictor that captures their shared context with experts tailored to each variable.

\noindent\textbf{Joint LLM-based Condition Prediction.}
We fine-tune a pretrained LLM to jointly predict temperature, time, and atmosphere from target and precursor compositions. This adapts the LLM's pretrained knowledge to synthesis-condition prediction and allows it to learn dependencies among the conditions. Since many synthesis records contain only a subset of these annotations, we train on all observed label combinations, including individual conditions and partially observed combinations, omitting missing fields from the output. This expands the usable training data beyond complete condition triples, enabling the model to learn a richer synthesis context.

\noindent\textbf{Variable-Specific Experts.}
The joint model's next-token prediction loss does not directly account for numerical error scales or atmosphere-class imbalance. We therefore train a separate gradient boosting machine (GBM)\citep{friedman2001greedy} for each variable using Magpie-based descriptors of the target and precursors\citep{ward2016general}. The temperature expert minimizes absolute error in degrees Celsius; the time expert does so in log space to account for its broad range. The atmosphere expert uses a class-balanced classification loss to account for label imbalance. These objectives provide variable-specific predictions that complement the joint model's shared synthesis context.

\noindent\textbf{Combining Joint and Variable-Specific Predictions.}
We combine the joint LLM's predictions with those of the variable-specific GBM experts. For each variable, we retain
the joint prediction when its absolute difference from the corresponding expert prediction is within a variable-specific threshold $\tau$ selected on validation data; otherwise, we average the two. Combination is performed on temperature in $^\circ$C, time in $\log(1+t)$, and atmosphere in probability space, yielding $\hat c_k=(\hat T_k,\hat t_k,\hat a_k)$. Implementation details are provided in Appendix~\ref{app:condition_combination_implementation}.

\subsection{Condition-Informed Synthesis Operation Prediction}
\label{Sec: Condition-Integrated Synthesis Operation Prediction}

The predicted conditions enter SOP through two complementary paths: they ground the LLM's generation and weight retrieved evidence from recorded procedures. We then fuse the candidates from both paths into a unified ranking.

\noindent\textbf{Condition-Grounded LLM Generation.}
We fine-tune the SOP model on target--precursor inputs augmented with condition descriptions from the frozen condition predictor, which provides the conditions at both training and inference. Conditioning on the processing regime introduces a synthesis-informed inductive bias: a high predicted temperature, for example, cues operations such as sintering. Since condition predictions are imperfect, the prompt describes the processing regime by coarse intervals, whereas retrieval and scheduling use the continuous values. Temperature is represented as $[\hat T_k-\Delta_{\mathrm{prompt}},\hat T_k+\Delta_{\mathrm{prompt}}]$, where $\Delta_{\mathrm{prompt}}$ is a fixed half-width, and time is assigned to a logarithmically spaced interval. Atmosphere is represented as \texttt{air} or \texttt{nonair} when available (e.g., $1120\,{}^\circ$C and 6 h become $1020$--$1220\,{}^\circ$C and $4$--$8$ h with $\Delta_{\mathrm{prompt}}=100\,{}^\circ$C; Appendix~\ref{app:retrieval_implementation}).

\noindent\textbf{Condition-Aware Retrieval.}
Alongside the LLM's knowledge, we retrieve evidence from recorded synthesis procedures. Shared precursors identify examples of how the same precursors are processed, so we score each candidate sequence by the number of precursors it shares with the query. However, precursor overlap alone treats procedures with different temperatures or times alike. To distinguish their processing regimes, we weight each procedure's contribution by a soft kernel over the predicted temperature and time that define its thermal regime. Let $\mathcal D_{\mathrm{train}}$ denote the database of synthesis procedures in the training set. For query $q=(m,p_k)$ and procedure $d\in\mathcal D_{\mathrm{train}}$ with precursor set $p_d$ and operation sequence $o_d$, we apply the same frozen condition predictor to both, giving a consistent basis for comparison without requiring complete condition annotations in the database. We define the weight\looseness-1
\begin{equation}
w(q,d)=
\exp\!\left(-\frac{|\hat T_k-\hat T_d|}{b_T}\right)
\exp\!\left(-\frac{|\log(1+\hat t_k)-\log(1+\hat t_d)|}{b_t}\right),
\label{eq:soft_condition_kernel}
\end{equation}
and score each operation sequence by
\begin{equation}
s_{\mathrm{Retrieval}}(o\mid m,p_k)=
\sum_{d\in\mathcal D_{\mathrm{train}}}
\mathbf 1[o_d=o]\,|p_k\cap p_d|\,w(q,d).
\label{eq:condition_retrieval_score}
\end{equation}
Here, $b_T,b_t>0$ control the temperature and log-time scales, and $|p_k\cap p_d|$ counts shared precursors. Since the predicted conditions are imperfect, the soft kernel gradually reduces the influence of dissimilar regimes while allowing procedures with differing conditions to contribute. The sum retains a frequency prior: sequences repeatedly observed for overlapping precursors and similar predicted conditions receive greater empirical support.  

\noindent\textbf{Candidate Fusion.}
To draw on complementary evidence from the LLM and recorded procedures, we fuse the generated and retrieved candidates into a single ranking. Because scores from the two sources differ in meaning and scale, we map each source's ranks to comparable scores:
$q_s(o)=\operatorname{Iso}_s(\operatorname{rank}_s(o))$.
Here, $\operatorname{Iso}_s$ is a non-increasing isotonic \citep{zadrozny2002transforming} fitted separately for each source to validation ranks and binary exact-match labels, ensuring that scores do not increase as rank worsens.
The score $q_s(o)$ estimates the exact-match frequency at the candidate's rank in source $s$. We set $q_s(o)=0$ for candidates absent from that source.
To preserve support from either source while rewarding agreement, we aggregate these scores using noisy-OR:
\begin{equation}
s_{\mathrm{fuse}}(o)=
1-\prod_{s\in\{\mathrm{Retrieval},\mathrm{LLM}\}}(1-q_s(o)).
\label{eq:candidate_fusion}
\end{equation}
The resulting score lies in $[0,1]$ and requires no additional mixing weight. The top $J$ distinct sequences form
$\mathcal O_{m,k}=(o_{k1},\ldots,o_{kJ})$.

\subsection{Resource-Aware Materials Synthesis Planning}
\label{sec:resource_planning}

Generating candidate routes leaves a downstream planning decision: which routes to attempt under limited experimental resources. A trial budget limits the number of routes attempted, while furnace-run and furnace-hour budgets account for equipment use. Predicted conditions specify the execution requirements used to assess sharing between thermal stages from different routes. We formulate this decision as a constrained optimization problem that jointly selects routes and allocates their thermal stages to maximize total utility under resource budgets.

\noindent{\textbf{Planning formulation.}}
For target $m$, let $\mathcal R_m$ contain routes $r_{kj}=(p_k,o_{kj})$ from the precursor candidates and their fused SOP lists. We select the route set $S$ that maximizes total utility within the imposed resource budgets:
\begin{equation}
\max_{S\subseteq\mathcal R_m}\ \sum_{r\in S}u(r)
\quad\mathrm{s.t.}\quad
C_q(S;\Delta_F)\leq B_q,\qquad q\in\mathcal Q,
\label{eq:resource_selection}
\end{equation}
where $S$ is the route set selected for target $m$, $u(r)$ is route utility and $C_q(S;\Delta_F)$ is the cost of selected routes under resource $q$, with budget $B_q$. Here, $\mathcal Q\subseteq\{\mathrm{trial},\mathrm{run},\mathrm{hour}\}$ refers to the specific type of imposed resource constraint, and $\Delta_F$ is the temperature tolerance for sharing a furnace run.

\noindent{\textbf{Route utility.}}
To compare candidate routes across precursor sets, we define route utility as the product of PP confidence and reciprocal rank in the corresponding fused SOP list:
\begin{equation}
u(r_{kj})=\frac{\pi_k^{\mathrm{PP}}}{j},
\qquad
\pi_k^{\mathrm{PP}}=
\sum_{z\in\mathcal Z_k}\exp\!\bigl(\ell(z\mid m)\bigr),
\label{eq:route_utility}
\end{equation}
where $\mathcal Z_k$ contains the distinct retained PP sequences representing $p_k$ after precursor normalization, and $\ell(z\mid m)$ sums the token log-probabilities of $z$ under the PP model. Summing over $\mathcal Z_k$ aggregates confidence across decoded representations of the same precursor set, while $1/j$ applies a common depth discount to each fused SOP list. Our resource-aware selection allocates the available
budget across precursor--SOP alternatives according to route utility,
balancing exploration across precursor sets with exploration of SOP
candidates within each set. Direct fusion-score weighting yields similar recovery in our comparisons (Appendix~\ref{app:fusion_score_utility}); we retain reciprocal rank because it uses the fused ordering without requiring comparable fusion-score gaps across precursor sets. Appendix~\ref{app:resource_planning_implementation} details sequence scoring and set-level aggregation.

\noindent{\textbf{Shared execution costs.}}
Each selected route uses a separate sample, so $C_{\mathrm{trial}}(S)=|S|$. Each heating, sintering, or annealing operation creates a thermal stage; non-thermal operations keep their order but add no furnace cost. Because stage-level annotations are incomplete, every thermal stage of $r_{kj}$ receives the predicted $(\hat T_k,\hat t_k,\hat a_k)$, charging the predicted maximum time at every stage (a repeated-maximum approximation). Since conditions are predicted per precursor set, they determine which precursor branches can share runs and how long each run occupies the furnace, while each route's operation sequence sets its number of thermal stages. Stages execute in rounds by position; those sharing a run must occupy the same position, have identical predicted times and atmosphere classes, and fit within capacity $C$, and a common setpoint must lie within $\Delta_F$ of every stage's predicted temperature. For fixed $S$, $C_{\mathrm{run}}(S;\Delta_F)$ is the minimum number of compatible runs; each run adds $\alpha+\hat t$ furnace-hours (overhead $\alpha$, shared time $\hat t$), giving $C_{\mathrm{hour}}(S;\Delta_F)$. Appendix~\ref{app:resource_planning_implementation} details compatibility, allocation, and stage-program sensitivity; $\Delta_F>0$ is evaluated in Appendix~\ref{app:temperature_milp}.\looseness-1

\noindent{\textbf{Joint selection and allocation.}}
With a trial budget alone, every route costs one trial, so selecting the highest-utility routes up to $B_{\mathrm{trial}}$ is optimal; this setting isolates the cross-branch utility of Eq.~\eqref{eq:route_utility}. With furnace constraints, a route's cost depends on the other selected routes, so a mixed-integer linear program (MILP) jointly selects routes and allocates all their thermal stages, admitting a route only if all its stages fit every imposed budget. Whether these plans recover reference recipes is measured separately. Appendix~\ref{app:resource_planning_implementation} gives the MILP, solver limits, and feasible-incumbent handling; Appendix~\ref{app:temperature_milp} compares greedy selection on the same pool and cost model.
\endgroup

\vspace{-0.7em}
\section{Experiments}
\vspace{-0.4em}
\label{sec: Experiments}
\begingroup
\setlength{\intextsep}{3pt}
\setlength{\textfloatsep}{5pt}
\setlength{\floatsep}{3pt}
\setlength{\columnsep}{11pt}
\captionsetup[table]{skip=3pt}
\newcommand{\MainSCPTable}{%
\begin{wraptable}{R}{0.48\textwidth}
\small
\setlength{\tabcolsep}{2.5pt}
\papertablesetup
\centering
\caption{Synthesis condition prediction.}
\label{tab:main_scp}
\resizebox{\linewidth}{!}{%
\begin{tabular}{lrrrrr}
\toprule
 & \multicolumn{2}{c}{Temperature} & \multicolumn{2}{c}{Time} & Atmosphere \\
\cmidrule(lr){2-3}\cmidrule(lr){4-5}\cmidrule(lr){6-6}
Model & MAE $\downarrow$ & RMSE $\downarrow$ & MAE $\downarrow$ & RMSE $\downarrow$ & Macro-F1 $\uparrow$ \\
\midrule
\citet{huo2022machine} & 151.76 & 195.29 & 13.11 & 25.64 & N/A \\
\addlinespace[0.8pt]
SyntMTE\citep{prein2025language} & \vsd{171.79}{8.71} & \vsd{217.60}{11.65} & N/A & N/A & N/A \\
\addlinespace[0.8pt]
\citet{karpovich2023interpretable} & \vsd{144.36}{0.23} & \vsd{188.85}{0.03} & \vsd{15.35}{0.25} & \vsd{25.31}{0.34} & N/A \\
\addlinespace[0.8pt]
GBM experts & 143.42 & 190.88 & 12.93 & 25.47 & 54.07 \\
\addlinespace[0.8pt]
Joint LLM & \vsd{138.68}{3.72} & \vsd{203.13}{1.82} & \vsd{13.92}{0.68} & \vsd{26.58}{1.22} & \vsd{63.56}{4.47} \\
\midrule
\proposed{} (mean) & \vsd{131.89}{3.18} & \vsd{\textbf{182.01}}{0.95} & \vsd{12.46}{0.28} & \vsd{\textbf{24.40}}{0.61} & \vsd{64.27}{0.10} \\
\addlinespace[0.8pt]
\proposed{} (threshold) & \vsd{\textbf{131.77}}{3.15} & \vsd{182.03}{0.87} & \vsd{\textbf{12.43}}{0.35} & \vsd{24.41}{0.69} & \vsd{\textbf{64.58}}{0.53} \\
\bottomrule
\end{tabular}}
\end{wraptable}%
}
\newcommand{\MainSOPTable}{%
\begin{wraptable}{r}{0.45\textwidth}
\small
\setlength{\tabcolsep}{2.5pt}
\papertablesetup
\centering
\caption{Synthesis operation prediction.}
\label{tab:main_sop}
\resizebox{\linewidth}{!}{%
\begin{tabular}{lrrrrrr}
\toprule
Method & Top-1 & Top-3 & Top-5 & Top-10 & Top-20 & Top-30 \\
\midrule
\rowcolor{gray!12}
\multicolumn{7}{l}{\textit{Specialized Models}} \\
\midrule
MPC\citep{he2023precursor} & \vsd{3.74}{0.38} & \vsd{9.69}{0.22} & \vsd{13.58}{0.82} & \vsd{19.46}{0.38} & \vsd{25.82}{0.66} & \vsd{32.39}{0.45} \\
\addlinespace[0.8pt]
SPENDE\citep{na2023artificial} & \vsd{6.68}{0.66} & N/A & N/A & N/A & N/A & N/A \\
\addlinespace[0.8pt]
MSP-LLM\citep{noh2026msp} & \vsd{10.71}{0.13} & \vsd{18.92}{0.23} & \vsd{22.88}{0.76} & \vsd{30.28}{0.19} & \vsd{40.23}{0.56} & \vsd{46.51}{0.29} \\
\midrule
\rowcolor{gray!12}
\multicolumn{7}{l}{\textit{Large Language Models}} \\
\midrule
GPT-5.4 mini (zero-shot) & \vsd{0.00}{0.00} & \vsd{0.05}{0.08} & \vsd{0.16}{0.23} & \vsd{2.72}{0.15} & \vsd{7.73}{1.08} & \vsd{11.00}{0.46} \\
\addlinespace[0.8pt]
GPT-5.4 mini (100-shot) & \vsd{11.06}{0.08} & \vsd{13.45}{0.08} & \vsd{15.20}{0.08} & \vsd{19.50}{0.31} & \vsd{24.89}{0.85} & \vsd{28.27}{0.85} \\
\addlinespace[0.8pt]
GPT-4o mini (fine-tuned) & \vsd{10.97}{0.13} & \vsd{11.55}{0.38} & \vsd{12.93}{0.33} & \vsd{17.79}{0.62} & \vsd{25.53}{0.77} & \vsd{28.32}{0.50} \\
\midrule
\makecell[l]{\proposed{}\\(w/o conditions)}
& \vsd{11.11}{0.00} & \vsd{19.14}{0.17}
& \vsd{24.36}{0.41} & \vsd{32.64}{0.44}
& \vsd{42.16}{0.50} & \vsd{47.02}{0.67} \\
\addlinespace[0.8pt]
\proposed & \vsd{\textbf{11.40}}{0.13} & \vsd{\textbf{20.12}}{0.74} & \vsd{\textbf{24.80}}{0.38} & \vsd{\textbf{34.24}}{1.09} & \vsd{\textbf{43.72}}{0.85} & \vsd{\textbf{49.06}}{0.49} \\
\bottomrule
\end{tabular}}
\end{wraptable}%
}

We evaluate \proposed{} along three questions: 
(1) whether the hybrid condition model improves synthesis-condition prediction (Section \ref{exp: scp}), 
(2) whether predicted conditions provide useful context for synthesis operation prediction (Section \ref{exp: sop}), 
and (3) whether resource-aware allocation improves recovery of literature-supported synthesis routes under limited experimental budgets (Section \ref{exp: msp}).
\proposed{} uses the same frozen PP model as MSP-LLM, isolating the effects of condition-informed operation prediction and route selection (PP accuracy in Appendix~\ref{app:shared_pp}).
\looseness-1

\noindent\textbf{Datasets.} Our main benchmark is curated from the text-mined inorganic synthesis recipes of \citet{kononova2019text}, following \citet{noh2026msp}. It is split into 8,295 training, 932 validation, and 918 test targets with no target shared across splits, and each test target has one reference precursor set and operation sequence. Condition prediction additionally uses 27,110 training records annotated with at least one of temperature, time, or atmosphere; only 11,277 (41.6\%) contain all three, which motivates training on all observed label combinations. We also construct a new evaluation dataset from the 80,806 solid-state synthesis records released by \citet{lee2025text}. After excluding targets present in the original splits or the condition-training corpus and curating the recorded operations, we obtain 2,344 new targets with 2,482 reference recipes, each pairing a precursor set with an operation sequence; 108 targets have multiple references, which differ in their operation sequences, precursor sets, or both, reflecting the one-to-many relationship between a target and its valid syntheses \citep{pan2026diffsyn}. Relative to the original training recipes, the new references contain far fewer oxide targets (32.6\% versus 92.5\%) and more all-elemental precursor sets (64.6\% versus 6.6\%). The pipeline trained on the main benchmark is applied without retraining, testing generalization beyond the chemistry that dominates its training data. More details are provided in Appendix~\ref{app:dataset}. \looseness=-1

\noindent\textbf{Implementation details.} Our trainable LLM components use LLaMA-3.1-8B-Instruct \citep{grattafiori2024llama} with QLoRA \citep{dettmers2023qlora, hu2021lora}. SOP and planning results report the mean $\pm$ sample SD over three seeds, except for condition-aware retrieval, which is deterministic. More details are provided in Appendix~\ref{app:implementation_details}. 

\subsection{Synthesis condition prediction (SCP)}
\label{exp: scp}
\vspace{-0.35em}
\MainSCPTable
\leavevmode{\textbf{Baseline methods.}} We compare against the predictor of \citet{huo2022machine}, SyntMTE \citep{prein2025language}, \citet{karpovich2023interpretable}, and our GBM experts without the joint LLM. Variables that a baseline does not predict are marked N/A. Adaptations and training-data variants are detailed in Appendices~\ref{app:baselines} and~\ref{app:scp_details}.
\newpage
\noindent\textbf{Metrics.} Given reference precursors, we evaluate maximum temperature and time using mean absolute error (MAE) and root mean squared error (RMSE). Atmosphere prediction uses macro-F1 over air/nonair.

\noindent\textbf{Results.}
Table~\ref{tab:main_scp} shows that combining the joint LLM with
variable-specific experts improves prediction over either component
alone. 
Both combination strategies outperform the individual predictors, suggesting that the joint LLM and variable-specific experts provide complementary predictive information.\looseness-1

\Needspace{4\baselineskip}
\subsection{Synthesis operation prediction (SOP)}
\label{exp: sop}
\MainSOPTable
\leavevmode{\textbf{Baseline methods.} The retrieval-based baseline MPC adapts the material-similarity
retriever of \citet{he2023precursor}. We also compare with SPENDE \citep{na2023artificial}.
LLM baselines include MSP-LLM \citep{noh2026msp}, fine-tuned GPT-4o mini,
and GPT-5.4 mini with zero-shot or 100-shot prompting.
MSP-LLM uses the same backbone and SOP training recipe as our generator,
without synthesis-condition inputs. \proposed{} (w/o conditions) combines MSP-LLM with our precursor-overlap retriever without condition weighting, retaining the same rank-calibration and fusion procedures. Appendix~\ref{app:baselines} specifies these adaptations and prompting settings.}

\noindent\textbf{Metrics.} Top-$k$ exact match is the fraction of targets whose reference operation sequence appears exactly, including order, among the top $k$ candidates. We report $k\in\{1,3,5,10,20,30\}$. All methods receive reference precursors, isolating operation prediction from precursor-selection errors. Supplementary NED similarity, LCS-F1, and multiset-F1 scores are defined and reported in Appendix~\ref{app:soft_metrics}.

\noindent\textbf{Results.}
Predicted conditions improve SOP coverage: they raise the rate at
which the reference sequence appears among the candidates, and the
improvement grows with candidate depth (Table~\ref{tab:main_sop}).
\proposed{} outperforms MSP-LLM at every reported $k$, and since the
two share the same backbone and SOP training recipe, the gap reflects
only what \proposed{} adds to generation. This gap is small at Top-1
(11.40\% vs.\ 10.71\%) but widens to 3.96 points at Top-10
(34.24\% vs.\ 30.28\%) and 2.55 points at Top-30 (49.06\% vs.\ 46.51\%).
Comparing with \proposed{} (w/o conditions) attributes this widening
to conditions rather than to retrieval and fusion alone.
Precursor-overlap retrieval and fusion account for most of the
Top-10 gain (30.28\% to 32.64\%) but contribute only 0.51 points at
Top-30, whereas conditions add 1.60 points at Top-10 and 2.04 points
at Top-30 (47.02\% to 49.06\%), making them the main source of
improvement at the depth downstream planning draws on. Since the conditions are themselves predicted from the target and precursors, this gain does not come from new inputs but from how the
processing regime is made explicit: the condition model draws on
condition-annotated records, including those unusable for SOP
training (Section~\ref{sec: Preliminaries}), and its predictions give
retrieval a continuous similarity over temperature and time where
precursor overlap alone treats all procedures sharing precursors
alike (paired confidence intervals in
Appendix~\ref{app:sop_paired_uncertainty}).

\subsection{Resource-aware materials synthesis planning}
\label{exp: msp}
\vspace{-0.35em}
\leavevmode{\textbf{Baseline methods.}}
We compare MSP-LLM, which explores SOP candidates for its Top-1
predicted precursor set, with fine-tuned GPT-4o mini and zero-shot/100-shot GPT-5.4 mini.
Each API-based model is evaluated both end-to-end and as a two-stage PP$\rightarrow$SOP pipeline.
The end-to-end variant directly generates precursor--SOP pairs; the two-stage variant first predicts one precursor set and then generates SOP candidates for it, following MSP-LLM's Top-1 strategy. All methods use predicted precursors throughout. \proposed{} performs utility-based route selection
using MILP, while baselines admit whole routes in their original
candidate order. To separate candidate construction from route selection, we also cross the two: MSP-LLM candidates (30 beams for each of the shared PP model's Top-10 precursor sets) with utility-based selection, and \proposed{} candidates admitted in order within the Top-1 precursor branch (Appendix~\ref{app:baselines}).\looseness-1

\noindent\textbf{Metrics.} For each target $m$, a method selects candidate routes $S$ under a per-target budget, counted in trials (one per selected route) or in furnace runs or furnace-hours. Under furnace budgets, the thermal stages of the selected routes are scheduled together, with predicted conditions determining compatibility and cost; the number of selected routes is then limited only by the furnace budget and the size of each method's candidate pool. All methods are scheduled with the same predicted conditions and simulated cost model (stages grouped by thermal-stage position, capacity $C=8$, 2\,h overhead per run, exact-temperature sharing with $\Delta_F=0$), so comparisons isolate candidate construction and route selection. Recipe recovery is the percentage of targets for which some route in $S$ exactly matches both the precursor set and the ordered operation sequence of a single reference recipe (any one reference suffices). Further details are provided in Appendices~\ref{app:evaluation_details} and \ref{app:resource_planning_implementation}.

\noindent\textbf{Results.} \proposed{} improves recipe recovery from 27.89\% to 32.90\% at three furnace runs and from 34.68\% to 40.01\% at 30 trials (Table~\ref{tab:main_msp_budgets}). At 30 trials, the crossed comparisons show gains from both candidate construction (34.68$\rightarrow$37.07\% under Top-1 admission; 37.22$\rightarrow$40.01\% under utility-based selection) and selection (34.68$\rightarrow$37.22\% with MSP-LLM candidates; 37.07$\rightarrow$40.01\% with \proposed{} candidates). As trial-only selection reduces to utility ranking, this selection gain comes from spreading trials across precursor sets: the Top-1 set misses the reference for 236 targets, and 186 of them have it in the top-10 (Appendix~\ref{app:shared_pp}). In the matched Top-10 setting, \proposed{} (w/o conditions) reaches 37.87\%, so conditions in candidate construction add 2.14 points (Appendix~\ref{app:planning_condition_control}). Under furnace budgets, utility and joint allocation both contribute: at 48 h, Top-1 admission of the same candidates recovers 29.05\%, utility-greedy 30.79\%, and MILP, which also accounts for shared runs, 31.99\% (Appendix~\ref{app:temperature_milp}). Some baselines lose recovery at larger furnace budgets because ordered admission is recomputed per budget (Appendix~\ref{app:furnace_nonmonotone}).\looseness-1

\providecommand{\vsd}[2]{\begin{tabular}[t]{@{}r@{}}#1\\[-2pt]{\scriptsize($\pm$\,#2)}\end{tabular}}
\providecolor{skyblue}{RGB}{206,229,250}
\begin{table}[tbp]
\centering
\caption{Material synthesis planning under trial and furnace budgets.}
\label{tab:main_msp_budgets}
\begingroup
\footnotesize
\setlength{\tabcolsep}{1.8pt}
\papertablesetup
\renewcommand{\arraystretch}{1.0}
\resizebox{0.88\linewidth}{!}{%
\begin{tabular}{l rrrrrr @{}p{8pt}@{}!{\color{gray!55}\vrule width 0.5pt}@{}p{8pt}@{} rrrrr @{\hspace{5pt}} rrrr}
\toprule
\multicolumn{1}{l}{\cellcolor{skyblue}} & \multicolumn{6}{c}{\cellcolor{skyblue}\textbf{(a) Trial budget} {\scriptsize(number of syntheses tried)}} & \cellcolor{skyblue} & \cellcolor{skyblue} & \multicolumn{9}{c}{\cellcolor{skyblue}\textbf{(b) Furnace budget} {\scriptsize(runs \,$|$\, hours)}} \\
\cmidrule(lr){2-7}\cmidrule(lr){10-14}\cmidrule(lr){15-18}
Method & 1 & 3 & 5 & 10 & 20 & 30 & & & 1 run & 2 runs & 3 runs & 5 runs & 10 runs & 12\,h & 24\,h & 48\,h & 96\,h \\
\midrule
\makecell[l]{GPT-5.4 mini\\(end-to-end, 100-shot)}
& \vsd{5.99}{0.31} & \vsd{8.71}{0.46} & \vsd{9.91}{0.31} & \vsd{12.20}{0.00} & \vsd{15.14}{0.31} & \vsd{16.72}{0.85}
& &
& \vsd{10.84}{0.39} & \vsd{11.66}{0.15} & \vsd{13.24}{0.54} & \vsd{14.87}{0.69} & \vsd{16.50}{0.69}
& \vsd{5.45}{0.15} & \vsd{9.91}{0.00} & \vsd{13.51}{0.31} & \vsd{15.09}{0.85} \\
\addlinespace[0.8pt]
\makecell[l]{GPT-5.4 mini\\(PP$\rightarrow$SOP, 100-shot)}
& \vsd{7.19}{0.31} & \vsd{8.61}{0.31} & \vsd{9.97}{0.54} & \vsd{12.69}{0.08} & \vsd{15.80}{0.31} & \vsd{18.08}{0.31}
& &
& \vsd{12.64}{0.77} & \vsd{12.31}{0.00} & \vsd{14.81}{0.00} & \vsd{17.27}{0.54} & \vsd{18.08}{0.31}
& \vsd{6.37}{0.39} & \vsd{11.22}{0.77} & \vsd{15.31}{0.08} & \vsd{17.27}{0.39} \\
\addlinespace[0.8pt]
\makecell[l]{GPT-4o mini\\(end-to-end, fine-tuned)}
& \vsd{7.55}{0.31} & \vsd{8.46}{0.67} & \vsd{10.20}{0.33} & \vsd{14.56}{0.77} & \vsd{20.26}{0.98} & \vsd{21.57}{1.04}
& &
& \vsd{11.73}{0.23} & \vsd{15.94}{1.06} & \vsd{19.39}{1.00} & \vsd{21.24}{1.00} & \vsd{21.53}{1.09}
& \vsd{6.64}{0.11} & \vsd{13.36}{0.38} & \vsd{18.23}{0.62} & \vsd{20.73}{1.14} \\
\addlinespace[0.8pt]
\makecell[l]{GPT-4o mini\\(PP$\rightarrow$SOP, fine-tuned)}
& \vsd{8.21}{0.25} & \vsd{8.79}{0.25} & \vsd{9.84}{0.33} & \vsd{13.65}{0.69} & \vsd{19.72}{0.58} & \vsd{21.60}{0.54}
& &
& \vsd{11.51}{1.11} & \vsd{15.98}{0.45} & \vsd{19.21}{0.74} & \vsd{21.53}{0.41} & \vsd{21.60}{0.54}
& \vsd{6.32}{0.11} & \vsd{12.82}{0.27} & \vsd{18.05}{0.35} & \vsd{20.73}{0.77} \\
\midrule
\rowcolor{gray!12}
\multicolumn{18}{l}{\textit{Candidate construction $\times$ route selection}} \\
\midrule
MSP-LLM
& \vsd{7.81}{0.13} & \vsd{13.94}{0.00} & \vsd{16.88}{0.39} & \vsd{22.55}{0.47} & \vsd{30.17}{0.39} & \vsd{34.68}{0.60}
& &
& \vsd{20.04}{0.11} & \vsd{21.57}{0.33} & \vsd{27.89}{0.71} & \vsd{32.75}{0.25} & \vsd{34.68}{0.60}
& \vsd{11.37}{0.17} & \vsd{20.04}{0.50} & \vsd{27.92}{0.44} & \vsd{32.61}{0.38} \\
\addlinespace[0.8pt]
\makecell[l]{MSP-LLM\\(w/ \proposed{} selection)}
& \vsd{7.81}{0.13} & \vsd{14.20}{0.23} & \vsd{17.97}{0.19} & \vsd{23.86}{0.68} & \vsd{31.55}{0.41} & \vsd{37.22}{0.25}
& &
& \vsd{20.23}{0.17} & \vsd{26.18}{0.44} & \vsd{30.72}{0.29} & \vsd{36.20}{0.49} & \vsd{40.01}{0.77}
& \vsd{11.44}{0.19} & \vsd{21.28}{0.67} & \vsd{30.76}{0.06} & \vsd{37.25}{0.11} \\
\addlinespace[0.8pt]
\makecell[l]{\proposed{}\\(w/ MSP-LLM selection)}
& \vsd{8.35}{0.23} & \vsd{14.52}{0.38} & \vsd{17.65}{0.11} & \vsd{25.05}{0.98} & \vsd{32.72}{0.60} & \vsd{37.07}{0.45}
& &
& \vsd{21.64}{0.55} & \vsd{23.06}{0.06} & \vsd{30.28}{0.76} & \vsd{35.95}{0.44} & \vsd{37.07}{0.45}
& \vsd{12.42}{0.11} & \vsd{21.75}{0.45} & \vsd{29.05}{0.38} & \vsd{34.68}{0.25} \\
\addlinespace[0.8pt]
\proposed{}
& \vsd{\textbf{8.46}}{0.23} & \vsd{\textbf{14.78}}{0.41} & \vsd{\textbf{18.81}}{0.17} & \vsd{\textbf{26.18}}{1.04} & \vsd{\textbf{34.42}}{0.11} & \vsd{\textbf{40.01}}{0.35}
& &
& \vsd{\textbf{21.71}}{0.56} & \vsd{\textbf{27.78}}{0.33} & \vsd{\textbf{32.90}}{0.47} & \vsd{\textbf{38.78}}{0.19} & \vsd{\textbf{42.85}}{0.38}
& \vsd{\textbf{12.49}}{0.06} & \vsd{\textbf{23.20}}{0.47} & \vsd{\textbf{31.99}}{0.38} & \vsd{\textbf{39.22}}{0.11} \\
\bottomrule
\end{tabular}}
\endgroup
\end{table}

\vspace{0.4em}

\providecommand{\vsd}[2]{\begin{tabular}[t]{@{}r@{}}#1\\[-2pt]{\scriptsize($\pm$\,#2)}\end{tabular}}
\providecolor{skyblue}{RGB}{206,229,250}
\begin{table}[!htbp]
\centering
\caption{Material synthesis planning on the new target set under trial and furnace budgets. Single and Multi denote single-reference ($n=2{,}236$) and multi-reference ($n=108$) targets.}
\label{tab:main_unseen_trials}
\begingroup
\footnotesize
\papertablesetup
\setlength{\tabcolsep}{1.2pt}
\renewcommand{\arraystretch}{1.0}
\resizebox{\linewidth}{!}{%
\begin{tabular}{@{}l
rrrrrr @{\hspace{4pt}} rrrrrr
@{\hspace{4pt}}!{\color{gray!55}\vrule width 0.5pt}@{\hspace{4pt}}
rrrrr @{\hspace{4pt}} rrrrr
@{\hspace{4pt}}!{\color{gray!55}\vrule width 0.5pt}@{\hspace{4pt}}
rrrr @{\hspace{4pt}} rrrr@{}}
\toprule
\cellcolor{skyblue}
& \multicolumn{12}{c}{\cellcolor{skyblue}\textbf{(a) Trial budget}}
& \multicolumn{18}{c}{\cellcolor{skyblue}\textbf{(b) Furnace budget}} \\
& \multicolumn{12}{c}{\scriptsize \textbf{Number of syntheses tried}}
& \multicolumn{10}{c}{\textbf{Runs}}
& \multicolumn{8}{c}{\textbf{Hours}} \\
\cmidrule(lr){2-13}\cmidrule(lr){14-23}\cmidrule(lr){24-31}
& \multicolumn{6}{c}{Single-ref.}
& \multicolumn{6}{c}{Multi-ref.}
& \multicolumn{5}{c}{Single-ref.}
& \multicolumn{5}{c}{Multi-ref.}
& \multicolumn{4}{c}{Single-ref.}
& \multicolumn{4}{c}{Multi-ref.} \\
\cmidrule(lr){2-7}\cmidrule(lr){8-13}
\cmidrule(lr){14-18}\cmidrule(lr){19-23}
\cmidrule(lr){24-27}\cmidrule(lr){28-31}
Method
& 1 & 3 & 5 & 10 & 20 & 30
& 1 & 3 & 5 & 10 & 20 & 30
& 1 & 2 & 3 & 5 & 10
& 1 & 2 & 3 & 5 & 10
& 12 & 24 & 48 & 96
& 12 & 24 & 48 & 96 \\
\midrule
\makecell[l]{GPT-5.4 mini\\(end-to-end, 100-shot)}
& \vsd{3.94}{0.19} & \vsd{5.64}{0.19} & \vsd{6.42}{0.22} & \vsd{7.87}{0.19} & \vsd{9.73}{0.28} & \vsd{10.73}{0.13}
& \vsd{\textbf{15.74}}{2.62} & \vsd{17.59}{3.93} & \vsd{20.37}{2.62} & \vsd{25.00}{2.62} & \vsd{25.93}{2.62} & \vsd{30.09}{1.96}
& \vsd{6.48}{0.38} & \vsd{7.76}{0.09} & \vsd{8.99}{0.06} & \vsd{10.00}{0.09} & \vsd{10.51}{0.25}
& \vsd{24.07}{0.00} & \vsd{26.85}{1.31} & \vsd{27.78}{1.31} & \vsd{28.24}{1.96} & \vsd{29.63}{1.31}
& \vsd{2.71}{0.35} & \vsd{5.50}{0.25} & \vsd{8.32}{0.06} & \vsd{9.97}{0.25}
& \vsd{6.48}{1.31} & \vsd{12.04}{2.62} & \vsd{20.83}{3.27} & \vsd{28.24}{1.96} \\
\addlinespace[0.8pt]
\makecell[l]{GPT-5.4 mini\\(PP$\rightarrow$SOP, 100-shot)}
& \vsd{4.18}{0.03} & \vsd{5.64}{0.06} & \vsd{7.09}{0.41} & \vsd{9.68}{0.35} & \vsd{12.61}{0.19} & \vsd{14.53}{0.38}
& \vsd{15.28}{0.65} & \vsd{18.98}{1.96} & \vsd{\textbf{27.31}}{1.96} & \vsd{30.56}{0.00} & \vsd{37.04}{1.31} & \vsd{40.74}{1.31}
& \vsd{8.70}{0.09} & \vsd{9.48}{0.06} & \vsd{12.14}{0.09} & \vsd{14.33}{0.28} & \vsd{14.53}{0.38}
& \vsd{25.46}{1.96} & \vsd{31.48}{1.31} & \vsd{34.72}{0.65} & \vsd{40.74}{1.31} & \vsd{40.74}{1.31}
& \vsd{3.18}{0.13} & \vsd{7.00}{0.16} & \vsd{10.64}{0.44} & \vsd{13.37}{0.57}
& \vsd{9.26}{0.00} & \vsd{14.81}{1.31} & \vsd{26.85}{0.00} & \vsd{36.57}{0.65} \\
\addlinespace[0.8pt]
\makecell[l]{GPT-4o mini\\(end-to-end, fine-tuned)}
& \vsd{4.01}{0.07} & \vsd{4.22}{0.11} & \vsd{4.76}{0.14} & \vsd{7.53}{0.07} & \vsd{13.07}{0.83} & \vsd{17.68}{1.52}
& \vsd{\textbf{15.74}}{0.00} & \vsd{16.05}{0.53} & \vsd{17.28}{0.53} & \vsd{20.06}{1.93} & \vsd{30.25}{3.85} & \vsd{37.65}{3.74}
& \vsd{5.38}{0.17} & \vsd{9.21}{0.27} & \vsd{14.52}{0.68} & \vsd{17.49}{1.43} & \vsd{17.68}{1.52}
& \vsd{19.14}{0.53} & \vsd{22.84}{1.07} & \vsd{33.95}{4.66} & \vsd{37.65}{3.74} & \vsd{37.65}{3.74}
& \vsd{2.50}{0.04} & \vsd{6.51}{0.31} & \vsd{11.55}{0.60} & \vsd{14.88}{0.74}
& \vsd{5.25}{1.41} & \vsd{13.58}{1.07} & \vsd{25.00}{1.85} & \vsd{33.64}{3.25} \\
\addlinespace[0.8pt]
\makecell[l]{GPT-4o mini\\(PP$\rightarrow$SOP, fine-tuned)}
& \vsd{3.52}{0.14} & \vsd{3.83}{0.17} & \vsd{4.28}{0.09} & \vsd{6.13}{0.19} & \vsd{11.05}{0.39} & \vsd{16.26}{0.44}
& \vsd{13.58}{0.53} & \vsd{13.89}{0.93} & \vsd{14.20}{1.41} & \vsd{18.21}{0.53} & \vsd{26.85}{0.93} & \vsd{37.65}{3.51}
& \vsd{4.43}{0.09} & \vsd{7.80}{0.19} & \vsd{13.06}{0.58} & \vsd{16.19}{0.43} & \vsd{16.26}{0.44}
& \vsd{16.67}{0.93} & \vsd{20.37}{1.60} & \vsd{33.64}{1.41} & \vsd{37.65}{3.51} & \vsd{37.65}{3.51}
& \vsd{2.47}{0.07} & \vsd{5.87}{0.30} & \vsd{10.42}{0.59} & \vsd{14.04}{0.50}
& \vsd{5.25}{1.07} & \vsd{13.58}{4.28} & \vsd{25.00}{1.85} & \vsd{34.88}{3.51} \\
\midrule
\rowcolor{gray!12}
\multicolumn{31}{l}{\textit{Candidate construction $\times$ route selection}} \\
\midrule
MSP-LLM
& \vsd{4.13}{0.21} & \vsd{7.48}{0.32} & \vsd{9.08}{0.12} & \vsd{11.82}{0.14} & \vsd{16.74}{0.75} & \vsd{20.63}{0.92}
& \vsd{13.27}{1.41} & \vsd{20.99}{1.07} & \vsd{24.07}{0.93} & \vsd{28.09}{0.53} & \vsd{35.19}{1.85} & \vsd{44.75}{1.41}
& \vsd{9.41}{0.22} & \vsd{13.36}{0.36} & \vsd{16.35}{0.91} & \vsd{20.01}{0.92} & \vsd{20.63}{0.92}
& \vsd{25.00}{0.93} & \vsd{29.32}{0.53} & \vsd{36.42}{2.83} & \vsd{43.21}{1.41} & \vsd{44.75}{1.41}
& \vsd{3.94}{0.15} & \vsd{9.35}{0.22} & \vsd{14.21}{0.42} & \vsd{17.96}{0.74}
& \vsd{7.10}{1.41} & \vsd{16.36}{1.93} & \vsd{28.09}{0.53} & \vsd{38.58}{1.07} \\
\addlinespace[0.8pt]
\makecell[l]{MSP-LLM\\(w/ \proposed{} selection)}
& \vsd{4.17}{0.21} & \vsd{\textbf{7.90}}{0.37} & \vsd{9.50}{0.14} & \vsd{12.21}{0.04} & \vsd{17.47}{0.81} & \vsd{21.50}{0.87}
& \vsd{13.27}{1.41} & \vsd{20.99}{1.07} & \vsd{24.07}{0.93} & \vsd{28.09}{0.53} & \vsd{35.19}{1.85} & \vsd{44.44}{1.85}
& \vsd{9.41}{0.22} & \vsd{14.15}{0.47} & \vsd{18.13}{0.93} & \vsd{21.00}{1.02} & \vsd{23.39}{0.91}
& \vsd{25.31}{0.53} & \vsd{29.63}{0.93} & \vsd{38.27}{1.41} & \vsd{43.21}{1.41} & \vsd{45.37}{2.45}
& \vsd{4.16}{0.19} & \vsd{9.97}{0.34} & \vsd{15.46}{0.66} & \vsd{19.57}{0.58}
& \vsd{8.02}{0.53} & \vsd{16.98}{1.41} & \vsd{28.70}{0.00} & \vsd{39.20}{3.25} \\
\addlinespace[0.8pt]
\makecell[l]{\proposed{}\\(w/ MSP-LLM selection)}
& \vsd{4.20}{0.37} & \vsd{7.27}{0.05} & \vsd{9.42}{0.26} & \vsd{13.86}{0.12} & \vsd{20.54}{0.70} & \vsd{26.45}{0.26}
& \vsd{14.51}{1.93} & \vsd{\textbf{22.22}}{0.93} & \vsd{25.31}{0.53} & \vsd{\textbf{30.86}}{0.53} & \vsd{37.96}{0.00} & \vsd{48.77}{1.41}
& \vsd{10.79}{0.17} & \vsd{15.85}{0.11} & \vsd{21.44}{0.34} & \vsd{25.94}{0.12} & \vsd{26.45}{0.26}
& \vsd{29.94}{0.53} & \vsd{33.02}{1.07} & \vsd{37.65}{0.53} & \vsd{47.84}{1.41} & \vsd{48.77}{1.41}
& \vsd{4.79}{0.13} & \vsd{11.75}{0.14} & \vsd{18.68}{0.14} & \vsd{23.24}{0.11}
& \vsd{11.11}{0.00} & \vsd{20.37}{0.00} & \vsd{34.88}{2.14} & \vsd{42.59}{0.93} \\
\addlinespace[0.8pt]
\proposed{}
& \vsd{\textbf{4.25}}{0.37} & \vsd{7.65}{0.04} & \vsd{\textbf{9.88}}{0.24} & \vsd{\textbf{14.04}}{0.15} & \vsd{\textbf{21.27}}{0.72} & \vsd{\textbf{27.55}}{0.31}
& \vsd{14.51}{1.93} & \vsd{\textbf{22.22}}{0.93} & \vsd{25.31}{0.53} & \vsd{\textbf{30.86}}{0.53} & \vsd{\textbf{38.27}}{0.53} & \vsd{\textbf{49.07}}{1.60}
& \vsd{\textbf{10.88}}{0.13} & \vsd{\textbf{17.11}}{0.36} & \vsd{\textbf{23.46}}{0.23} & \vsd{\textbf{27.24}}{0.09} & \vsd{\textbf{29.59}}{0.27}
& \vsd{\textbf{30.25}}{0.53} & \vsd{\textbf{34.26}}{0.93} & \vsd{\textbf{41.36}}{0.53} & \vsd{\textbf{49.69}}{1.41} & \vsd{\textbf{50.00}}{0.93}
& \vsd{\textbf{5.11}}{0.14} & \vsd{\textbf{12.28}}{0.10} & \vsd{\textbf{19.74}}{0.07} & \vsd{\textbf{25.61}}{0.23}
& \vsd{\textbf{11.42}}{0.53} & \vsd{\textbf{21.60}}{0.53} & \vsd{\textbf{35.80}}{2.14} & \vsd{\textbf{45.99}}{1.41} \\
\bottomrule
\end{tabular}}
\endgroup
\end{table}

\noindent\textbf{Generalization under distribution shift.} The new dataset \citep{lee2025text} departs substantially from the training distribution, with far fewer oxide targets and many more all-elemental precursor sets. Applied without retraining, \proposed{} achieves 30-trial recovery of 27.55\% on single-reference targets and 49.07\% on multi-reference targets, compared with 20.63\% and 44.75\% for MSP-LLM (Table~\ref{tab:main_unseen_trials}). Much of this improvement comes from \proposed{}'s candidates: with MSP-LLM selection, they already reach 26.45\% and 48.77\%, and utility-based selection adds further gains, particularly under furnace budgets. The advantage also holds in both oxide and non-oxide subsets (Appendix~\ref{app:new_target_results}), supporting generalization beyond the oxide-dominated training chemistry. Zero-shot results for both datasets are reported in Appendix~\ref{app:zero_shot_planning}.\looseness-1

\vspace{-0.4em}
\subsection{Ablation studies}
\vspace{-0.35em}

\providecommand{\abcmark}{\textcolor{blue}{\ding{51}}}
\providecommand{\abxmark}{\textcolor{red}{\ding{55}}}

\begin{wraptable}{R}{0.50\textwidth}
\centering
\caption{Ablation of \proposed{} components.}
\label{tab:main_ablation}
\begingroup
\footnotesize
\setlength{\tabcolsep}{4pt}
\papertablesetup
\renewcommand{\arraystretch}{1.0}

\resizebox{\linewidth}{!}{%
\begin{minipage}{\textwidth}
\centering

\begin{tabular*}{\linewidth}{@{\extracolsep{\fill}}lcc rr rr r}
\toprule
\rowcolor{gray!12}
\multicolumn{8}{l}{\textit{(a) Synthesis Condition Prediction (SCP)}} \\
\midrule
& & & \multicolumn{2}{c}{Temperature}
& \multicolumn{2}{c}{Time} & Atmosphere \\
\cmidrule(lr){4-5}\cmidrule(lr){6-7}\cmidrule(lr){8-8}
Model & Joint & GBM
& MAE $\downarrow$ & RMSE $\downarrow$
& MAE $\downarrow$ & RMSE $\downarrow$
& Macro-F1 $\uparrow$ \\
\midrule
(A) Separate LLMs
& \abxmark & \abxmark
& 141.42 & 207.30 & 14.01 & 26.33 & 60.90 \\
(B) Separate LLMs + GBM
& \abxmark & \abcmark
& 132.03 & 186.17 & 12.48 & 24.72 & 62.46 \\
(C) GBM experts only
& \abxmark & \abcmark
& 143.42 & 190.88 & 12.93 & 25.47 & 54.07 \\
(D) Joint LLM
& \abcmark & \abxmark
& 138.68 & 203.13 & 13.92 & 26.58 & 63.56 \\
\midrule
(E) \proposed{}
& \abcmark & \abcmark
& \textbf{131.77} & \textbf{182.03}
& \textbf{12.43} & \textbf{24.41} & \textbf{64.58} \\
\bottomrule
\end{tabular*}

\par\vspace{1pt}

\resizebox{\linewidth}{!}{%
\begin{tabular}{lccc rrr}
\toprule
\rowcolor{gray!12}
\multicolumn{7}{l}{\textit{(b) Synthesis Operation Prediction (SOP)}} \\
\midrule
Model & Conditions & Generation & Retrieval & Top-10 & Top-20 & Top-30 \\
\midrule
(A) MSP-LLM
& \abxmark & \abcmark & \abxmark
& 30.28 & 40.23 & 46.51 \\
(B) \proposed~generation
& \abcmark & \abcmark & \abxmark
& 32.03 & 41.18 & 47.89 \\
(C) \proposed~retrieval (overlap only)
& \abxmark & \abxmark & \abcmark
& 31.37 & 40.31 & 46.19 \\
(D) \proposed~retrieval
& \abcmark & \abxmark & \abcmark
& 34.20 & 41.94 & 47.82 \\
\midrule
(E) \proposed{}
& \abcmark & \abcmark & \abcmark
& \textbf{34.24} & \textbf{43.72} & \textbf{49.06} \\
\bottomrule
\end{tabular}%
}

\par\vspace{1pt}

\begin{tabular*}{\linewidth}{@{\extracolsep{\fill}}lc rrr rr rr}
\toprule
\rowcolor{gray!12}
\multicolumn{9}{l}{\textit{(c) Route selection on the same candidate pool}
($\Delta_F=0$)} \\
\midrule
& & \multicolumn{3}{c}{Trials}
& \multicolumn{2}{c}{Furnace runs}
& \multicolumn{2}{c}{Furnace hours} \\
\cmidrule(lr){3-5}\cmidrule(lr){6-7}\cmidrule(lr){8-9}
Method & Utility & 5 & 10 & 30 & 1 & 5 & 24 & 96 \\
\midrule
(A) Order by PP rank & \abxmark
& 17.65 & 25.05 & 37.07
& 21.64 & 35.98 & 22.19 & 36.82 \\
(B) Order by SOP rank & \abxmark
& 12.85 & 15.80 & 25.89
& 19.90 & 29.23 & 19.75 & 31.59 \\
\midrule
(C) \proposed{} & \abcmark
& \textbf{18.81} & \textbf{26.18} & \textbf{40.01}
& \textbf{21.71} & \textbf{38.78}
& \textbf{23.20} & \textbf{39.22} \\
\bottomrule
\end{tabular*}

\end{minipage}%
}

\endgroup
\end{wraptable}
\leavevmode
\noindent\textbf{Condition prediction.}
Table~\ref{tab:main_ablation}(a) examines joint prediction and
expert combination separately.
Adding variable-specific GBM experts improves both separate and
joint LLM predictors.
Combining the joint LLM with these experts achieves the best reported
mean scores, supporting their complementary roles in condition prediction.

\noindent\textbf{Operation prediction.} Table~\ref{tab:main_ablation}(b) uses reference precursors on all 918 targets to isolate condition use in generation and retrieval. Relative to MSP-LLM trained and evaluated without conditions, condition-grounded generation improves Top-10/30 exact match from 30.28/46.51\% to 32.03/47.89\%. Adding condition weighting to our precursor-overlap retriever improves these scores from 31.37/46.19\% to 34.20/47.82\%. Fusing the two sources gives the highest mean at every $k$, with the clearest gains at Top-20 and Top-30; the Top-10 mean is nearly tied with retrieval alone.\looseness-1

\noindent\textbf{Route selection.} Table~\ref{tab:main_ablation}(c) compares selection strategies on the same full \proposed{} candidate pool with predicted precursors. Ordering by PP rank exhausts the SOP candidates of each precursor set before moving to the next, whereas ordering by SOP rank takes candidates of the same SOP rank across precursor sets before moving to the next rank. At 30 trials, \proposed{} reaches 40.01\% recovery, versus 37.07\% and 25.89\% for these orderings, and it also achieves the highest recovery under every reported furnace budget. The trial-budget gain thus comes not from mere diversification but from weighting precursor sets by PP confidence.\looseness-1

\subsection{Further analyses}

\begin{table}[!htbp]
\begingroup
\footnotesize
\setlength{\tabcolsep}{2.5pt}
\papertablesetup
\begin{minipage}[t]{0.55\linewidth}
\centering
\caption{SOP generation with different condition sources at training and inference, on the subset with recorded conditions.}
\label{tab:main_condition_source}
\resizebox{\linewidth}{!}{%
\begin{tabular}{llrrr}
\toprule
Training & Inference & Top-10 & Top-20 & Top-30 \\
\midrule
None & None & \msd{24.02}{0.11} & \msd{33.06}{0.75} & \msd{38.61}{0.43} \\
Predicted & Predicted & \msd{25.08}{0.09} & \msd{34.94}{0.43} & \msd{40.21}{0.61} \\
Predicted & Ground truth & \msd{26.13}{0.71} & \msd{35.84}{0.43} & \msd{41.49}{0.53} \\
Ground truth & Ground truth & \msd{\textbf{27.11}}{0.85} & \msd{\textbf{36.82}}{0.75} & \msd{\textbf{42.92}}{0.70} \\
\bottomrule
\end{tabular}}
\end{minipage}\hfill
\begin{minipage}[t]{0.44\linewidth}
\centering
\caption{Top-30 SOP exact match (\%) by reference-sequence training frequency. Parentheses show each group's share of test targets.}
\label{tab:main_frequency}
\resizebox{\linewidth}{!}{%
\begin{tabular}{lrrr}
\toprule
 & Low {\scriptsize(28.1\%)} & Medium {\scriptsize(35.7\%)} & High {\scriptsize(36.2\%)} \\
\midrule
MSP-LLM & \msd{1.68}{0.59} & \msd{38.01}{0.70} & \msd{89.76}{0.80} \\
\proposed~retrieval & 1.55 & 33.84 & 97.59 \\
\proposed~generation & \msd{1.42}{1.12} & \msd{\textbf{42.28}}{1.07} & \msd{89.56}{0.63} \\
\midrule
\proposed{} & \msd{\textbf{2.07}}{0.22} & \msd{38.92}{1.98} & \msd{95.58}{0.46} \\
\bottomrule
\end{tabular}}
\end{minipage}
\endgroup
\end{table}

\noindent\textbf{Effect of condition source on generation.} Table~\ref{tab:main_condition_source} varies the source of conditions at training and inference. Because recorded conditions exist for only part of the data, all rows retrain the generator on 6,203 training recipes and evaluate on the 664 test targets with recorded conditions, using reference precursors without retrieval, so absolute scores are not comparable to Table~\ref{tab:main_sop}. Exact match increases monotonically from no conditions to predicted to ground-truth conditions at every $k$: predicted conditions raise Top-30 from 38.61\% to 40.21\%, and ground-truth conditions at inference alone add a further 1.28 points, reaching 42.92\% when also used in training. Since both sources share the same interval encoding, the generator is responding to the accuracy of the described processing regime rather than to the presence of a condition block, and predicted conditions already recover about 40\% of the oracle gain. Predicted conditions thus act as process context for operation prediction, and better condition prediction translates directly into better operation prediction (inference-time interventions in Appendix~\ref{app:condition_intervention}).\looseness-1

\noindent\textbf{Complementarity across sequence frequencies.}
Table~\ref{tab:main_frequency} reports Top-30 exact match with reference precursors, grouping targets by how often their reference sequence appears among the training recipes (Low: 0--4, Medium: 5--79, High: $\geq80$). Generation and retrieval are strong in different frequency ranges. Condition-grounded generation shows its largest gain on medium-frequency sequences, improving from 38.01\% for MSP-LLM to 42.28\%, whereas condition-aware retrieval performs best on frequent ones (97.59\%). Fusing both, \proposed{} exceeds MSP-LLM in every group and is best on rare sequences, although absolute recovery there remains low; it does not fully match the stronger source in the medium- and high-frequency groups, leaving room for improved fusion.

\FloatBarrier
\endgroup

\vspace{-0.7em}
\section{Conclusion}
\vspace{-0.4em}
\label{sec: Conclusion}

\begingroup
This paper presents \proposed{}, which extends MSP from predicting recipes to planning their execution by using predicted synthesis conditions both as process context for operation generation and retrieval and as execution requirements for route selection under trial and furnace budgets. \proposed{} reaches 49.06\% Top-30 exact match on SOP and raises reference-recipe recovery over MSP-LLM from 27.89\% to 32.90\% at three simulated furnace runs per target and from 34.68\% to 40.01\% at 30 trials, and its advantage persists without retraining on 2,344 new targets with a shifted chemistry distribution. The resulting per-target planning step provides a basis on which multi-target and closed-loop synthesis planners can build.
\endgroup
\par\FloatBarrier
\endgroup

\clearpage

\section*{Ethics Statement}
This work uses only publicly available, text-mined literature data
\citep{kononova2019text,lee2025text} and involves no human subjects,
personally identifiable information, or animal experiments. We are not aware
of conflicts of interest or sponsorship concerns affecting this work.

The plans \proposed{} produces are computational hypotheses derived from
reported syntheses, not validated procedures. Recipe recovery measures
agreement with literature references rather than experimental success, and
the furnace costs we report come from a simulated accounting model rather
than from laboratory execution (Appendix~\ref{app:limitations}). Predicted
conditions are point estimates with nonzero error, and inorganic synthesis
involves heating, controlled atmospheres, and reactive powders whose safe
handling depends on the specific chemistry. Any plan produced by \proposed{}
should therefore be reviewed by a qualified experimentalist, and checked
against the relevant safety and equipment constraints, before execution;
\proposed{} is intended to prioritize candidate routes for human or automated
laboratories under expert supervision, not to replace that judgment. 

\section*{Reproducibility Statement}
Source code for \proposed{}, including the condition model, the
condition-grounded generator and condition-aware retriever, the fusion
procedure, and the route-selection solver, is available at
\url{https://github.com/HeewoongNoh/CondMSP-Official}.
Section~\ref{sec: Methodology} defines the method, and
Appendix~\ref{app:implementation_details} gives training, decoding, retrieval,
evaluation, and resource-accounting settings, with the planning formulation and
solver configuration in Appendix~\ref{app:resource_planning_implementation}.
Appendix~\ref{app:dataset} documents both datasets, including the splits of the
main benchmark and the construction, curation, and verification of the new
evaluation set; Appendix~\ref{app:baselines} records how each baseline was
adapted and prompted, and Appendix~\ref{app:prompt_templates} lists the exact
model input and output formats. All reported numbers are means and sample
standard deviations over the seeds stated in
Appendix~\ref{app:training_details}, and the uncertainty procedures behind the
reported intervals are described in
Appendices~\ref{app:sop_paired_uncertainty} and~\ref{app:planning_condition_control}.

\section*{AI Use Statement}
In this work, we used generative AI tools to implement methods and to clean
and reformat a dataset: parts of the implementation and of the reproduced
baselines were written with the assistance of an LLM coding assistant, and an
LLM was used to curate operation sequences when constructing our new
evaluation set from the released records of \citet{lee2025text}
(Appendix~\ref{app:unseen_curation}). We have not used generative AI tools to
generate synthetic datasets, to develop theoretical models or conceptual
frameworks, to formulate mathematical claims, to provide critical ingredients
for proving mathematical claims, to assist in the writing of proofs, to propose
or refine the hypotheses of this work, to support qualitative or thematic data
analysis, or to interpret the reported results. Assistance with translation is
not applicable to this work. We also used generative AI tools to create and edit software code, to
create and modify figures, and to edit the manuscript for readability, structure, and layout.

We have reviewed all AI-assisted work. All AI-assisted code was read and tested
by the authors, and every number reported in this paper was produced by
executing that code on the stated data rather than generated by a language
model. The LLM-assisted dataset curation required agreement between two independent
responses, followed by a source-fidelity review that rejected invented or
reordered steps, and programmatic checks of event provenance and dataset
integrity; the protocol and its residual limitations are described in
Appendices~\ref{app:unseen_curation} and~\ref{app:limitations}. AI-assisted
figures were checked against the underlying result files. All claims in this
paper were formulated by the authors; generative AI tools were used only to
revise wording, formatting, and layout of text the authors had already written.
We take responsibility for the final content of this work, including text,
claims, or artifacts produced with the aid of generative AI.

We note separately that GPT-4o mini and GPT-5.4 mini appear as evaluated
baselines (Appendix~\ref{app:baselines}), and that one condition-prediction
comparison uses training data that the original authors generated with an LLM
(Appendix~\ref{app:scp_details}); these are part of the evaluated setting
rather than of the authoring process.

\bibliography{iclr2027_conference}
\bibliographystyle{iclr2027_conference}

\clearpage
\begin{center}
    \huge{\emph{Supplementary Material}}
\end{center}
\begin{center}
    \large{\emph{- CondMSP: Condition-Informed and Resource-Aware Inorganic Materials Synthesis Planning -}}
\end{center}
\vspace*{2mm}

{\small\DoToC}

\clearpage

\appendix
\renewcommand{\floatpagefraction}{0.8}
\makeatletter
\setlength{\@fptop}{0pt}
\setlength{\@fpsep}{12pt}
\setlength{\@fpbot}{0pt plus 1fil}
\makeatother
\begingroup
\section{Related Works}
\label{app:extended_related_works}

\paragraph{Materials Synthesis Prediction.}
In solid-state synthesis, the phases that form depend on precursor choice,
temperature, dwell time, and atmosphere
\citep{west2022solid,rao1997new,schubert2019synthesis}, and route design
has long relied on empirical knowledge despite early discussions of rational
design and planning of solid-state syntheses \citep{stein1993turning,jansen2002concept}.
Thermodynamics-guided approaches rank candidate reactions by computed
reaction energies and nucleation barriers \citep{aykol2021rational} or
search reaction networks built from computed phase stability
\citep{mcdermott2021graph}, but do not output literature-style recipes with
operation sequences and processing conditions. Text mining of the
synthesis literature has produced large recipe collections
\citep{kim2017materials,kononova2019text,lee2025text} that enable
data-driven prediction of such recipe components. Precursor recommendation has been addressed by materials-similarity
retrieval \citep{he2023precursor}, retrieval that combines expert
knowledge with attention over retrieved precursor sets
\citep{noh2024retrieval}, ranking models \citep{prein2025retro},
elementwise template formulation \citep{kim2024predicting}, and LLMs
\citep{kim2024large}, and synthesis-condition prediction by feature-based
regressors, reaction graph networks, and data-augmented language models
\citep{huo2022machine,karpovich2023interpretable,prein2024reaction,prein2025language}.
Recent work explores reinforcement learning with experimental feedback for
precursor reasoning and augmentation strategies for LLM-based synthesis
prediction \citep{pan2026synreason,prein2026benchmarking}, while DiffSyn
models distributions of synthesis compositions and conditions for zeolites
\citep{pan2026diffsyn}. For synthesis procedures, graph-based
representations of inorganic synthesis reactions have been proposed
\citep{andrello2025representation}, SPENDE predicts operations and associated
conditions for thermoelectric materials \citep{na2023artificial}, and
MSP-LLM connects precursor prediction with precursor-conditioned operation
generation \citep{noh2026msp}. \proposed{} differs in using predicted
conditions as process context for operation generation and as a basis for
weighting evidence from recorded procedures, combining both sources into a
unified candidate ranking.

\paragraph{Self-Driving Laboratories for Materials Synthesis.}
Self-driving laboratories integrate predictive models, automated
experiments, and feedback to guide synthesis
\citep{abolhasani2023rise,szymanski2023autonomous,hickman2025atlas}, and
recent reviews frame autonomous realization as the closing step of
AI-driven materials discovery \citep{lee2026toward,choi2026ai}. In such
platforms, samples are processed in batches, so throughput depends on
which experiments can share equipment. \proposed{} addresses the
per-target step of this pipeline: it formulates the selection of
precursor--operation alternatives under trial and furnace budgets, using
predicted conditions to determine which thermal stages can share furnace
capacity.
\endgroup
\begingroup
\section{Implementation Details}
\label{app:implementation_details}

\subsection{LLM training and decoding}
\label{app:training_details}
The precursor, joint-condition, and operation models use
LLaMA-3.1-8B-Instruct with 4-bit quantization and LoRA
\citep{grattafiori2024llama,dettmers2023qlora}.
Adapters use rank 16, scaling 32, dropout 0.1, and the query, key,
value, and output projections. All models train for three epochs with
cosine learning-rate decay, a 0.03 warmup ratio, and loss on assistant
response tokens only. The PP model uses learning rate $10^{-4}$ and
effective batch 8; the joint-condition model uses $10^{-4}$ and batch 4;
SOP uses $2\times10^{-4}$ and batch 2.
SOP runs use matched training seeds 42, 46, and 51.
We first train the condition predictor, then freeze it to generate condition inputs for SOP training and inference.

PP generates 10 beams. The joint-condition model generates five beams,
from which the highest-ranked output supplies the deployed condition fields.
SOP generates 30 beams with at most 256 new tokens and stops at the
completed output list. Sampling is disabled. Raw SOP beams are not
deduplicated for standalone evaluation; fusion merges identical operation
sequences across its two sources. Inputs contain the target and precursors,
with no article title or extracted synthesis-context field.
The SOP input also contains precursor-type tags derived from the precursor
formulas. Condition and SOP templates appear in Appendix~\ref{app:prompt_templates}.

\paragraph{Sensitivity to beam deduplication.}
Order-preserving deduplication of the saved 30-beam SOP outputs,
without replenishing candidates, changes mean exact match across the
reported $k$ by at most 0.04 percentage points for MSP-LLM and 0.07
for the \proposed{} generator. Top-10 remains 30.28\% for MSP-LLM
and changes from 32.03\% to 32.06\% for our generator, indicating
negligible sensitivity in these standalone SOP comparisons.

\subsection{Joint condition output and supervision}
\label{app:joint_condition_implementation}
The deployed joint-condition model is trained on 27,110 examples and emits
available fields in the order \texttt{temp}, \texttt{time}, \texttt{atmos}.
Missing training annotations are omitted from the response.
Temperature and time supervision use the maximum recorded
values. The joint LLM uses six atmosphere labels (air, nitrogen, oxygen,
hydrogen, argon, other), mapped to air/nonair downstream.

\subsection{Composition experts and prediction combination}
\label{app:condition_expert_implementation}
\label{app:condition_combination_implementation}
Let $\phi$ be the Magpie-based composition descriptor and
$\bar\phi(p)=|p|^{-1}\sum_{v\in p}\phi(v)$ the mean precursor descriptor.
Each expert receives
\begin{equation}
h(m,p)=[\phi(m);\bar\phi(p);\bar\phi(p)-\phi(m)].
\label{eq:expert_composition_features}
\end{equation}
The numerical gradient boosting machines use absolute-error loss,
300 iterations, and learning rate 0.06; temperature is fitted in
$^\circ$C and time in $\log(1+t)$, with $t$ in hours.
The atmosphere expert uses 400 iterations, the same learning rate,
and class-balanced binary classification. Each expert is fitted on
training rows with the corresponding annotation.

For a joint prediction $x$ and the corresponding expert prediction $y$, the variable-specific threshold rule is
{
\begin{equation}
G_{\tau}(x,y)=
\begin{cases}
x, & |x-y|\leq\tau,\\
(x+y)/2, & \text{otherwise}.
\end{cases}
\label{eq:condition_agreement}
\end{equation}
}
Temperature is combined directly using Eq.~\eqref{eq:condition_agreement}.
For time, define $\ell_\epsilon(t)=\log(1+\max(t,\epsilon))$,
with $\epsilon=0.05$ h. The combined prediction is
\begin{equation}
\hat t=\exp\!\left[G_{\tau_t}\bigl(
\ell_\epsilon(t_{\mathrm{LLM}}),\ell_\epsilon(t_{\mathrm{GBM}})
\bigr)\right]-1.
\label{eq:duration_combination}
\end{equation}
Atmosphere combines the LLM's nonair indicator with the expert's nonair
probability, then thresholds the result.
Validation selection gives $\tau_T=20\,{}^\circ$C,
$\tau_t=0.4$, $\tau_a=0.3$, and atmosphere threshold $\eta=0.4$.
Numerical gates minimize validation MAE; the atmosphere gate uses macro-F1.
If one numerical predictor omits a value, the other supplies it.
Atmosphere remains absent from the SOP prompt when the joint LLM omits it.

\subsection{Condition encoding, retrieval, and fusion}
\label{app:retrieval_implementation}
The SOP condition block uses a temperature interval of
$\hat T\pm100\,{}^\circ$C and time bins with boundaries
$0,0.5,1,2,4,8,16,32,64,128$ h (the last bin is open-ended).
Available atmosphere labels are air/nonair.
This encoding is fixed during training and inference and across
furnace-sharing scenarios. These intervals are prompt settings, not
calibrated confidence intervals; retrieval and scheduling use continuous
predicted values.

Retrieval uses the 8,295 training procedures, with conditions predicted
by the same frozen model used for queries, including procedures with
incomplete observed conditions.
A precursor-to-procedure lookup computes
Eq.~\eqref{eq:condition_retrieval_score} by visiting procedures that
share at least one precursor. The kernel scales are
$b_T=200\,{}^\circ$C and $b_t=1$ in log-time space;
atmosphere is omitted from the retrieval kernel. Contributions are summed
over training procedures sharing precursors with the query. For each SOP seed, isotonic rank calibration
is fitted on 932 validation queries with reference precursors and
applied unchanged to deployment queries with predicted precursors.
The union contains at most 30 distinct candidates from each source;
fusion returns its top 30 sequences.
Noisy-OR supplies a monotone ranking score; it is not treated as a calibrated
success probability and does not impose source independence.

\subsection{Evaluation}
\label{app:evaluation_details}
SCP methods are compared on the same evaluation rows for each variable.
SOP and planning use all 918 test representatives and report the mean
and sample SD across the three SOP seeds, except where stated otherwise.
API request variation is distinguished from training-seed variation
in Appendix~\ref{app:baselines}.

Target-paired confidence intervals resample targets after averaging the
matched per-target seed differences. They quantify target sampling
uncertainty, separately from the displayed seed SD.
Standalone SOP supplies the reference precursor set; full planning supplies
predicted precursors to condition prediction, generation, and retrieval.
The new target set uses paired references: both the precursor set and
operation sequence must match the same reference.

\paragraph{Condition interventions.}
Table~\ref{tab:app_condition_intervention} evaluates SOP generation alone
on all 918 test targets with reference precursors, using the same frozen
condition-trained SOP checkpoints (seeds 42, 46, and 51).
Only the inference condition input changes: we retain the original predicted
conditions, remove the condition block, or shuffle predicted condition blocks
across targets while preserving atmosphere-field presence.
The target, precursor input, and checkpoint remain fixed; no model is retrained,
and retrieval and fusion are not used. No recorded condition labels are required.
We report Top-10 and Top-30 operation-sequence exact match, as the mean and
sample SD across the three seeds. Unlike the separately trained configurations
in Table~\ref{tab:main_ablation}(b), this comparison tests sensitivity to the
condition context at inference.
\paragraph{Condition source in generation.}
Table~\ref{tab:main_condition_source} varies the condition source at
training and at inference. Because recorded conditions are available for only part of the data, all
four configurations use the same 6,203 training recipes and the same 664
test targets whose references carry recorded temperature and time;
atmosphere is included when recorded and omitted otherwise.
The subset is fixed before any model is trained. Each configuration fine-tunes the SOP
generator from scratch with the indicated training input (no condition
block, predicted conditions from the frozen condition model, or recorded
conditions) and is evaluated with reference precursors, without retrieval
or fusion, using the indicated inference input. Predicted conditions use
the same interval encoding as the main pipeline; recorded conditions are
encoded with the same format. We report Top-10, Top-20, and Top-30
exact match as mean and sample SD across three seeds.
\begingroup
\subsection{Resource-aware planning implementation}
\label{app:resource_planning_implementation}

\paragraph{Candidate scope and utility.}
Planning is performed independently for each target, using all distinct
precursor sets from the PP top-10 outputs and up to 30 fused SOP candidates
per set. Conditions, generation, and retrieval use each predicted
precursor, with no reference-precursor or top-precursor fallback.
In Eq.~\eqref{eq:route_utility}, $\mathcal Z_k$ groups the distinct retained
PP sequences representing the same unordered set after lowercasing and
removing spaces. The sequence score sums candidate-token log probabilities
without length normalization; set masses are aggregated by log-sum-exp,
counting identical serialized outputs once. Ordering uses
$\log\pi_k^{\mathrm{PP}}-\log j$, with ties broken by precursor rank and
then SOP rank. Normalizing PP masses within a target leaves both the
ordering and additive-utility maximizers unchanged.

\paragraph{Predicted thermal programs.}
Each route is a separate sample; intermediate products are not shared.
Every heating, sintering, or annealing operation contributes one of its
$H_r$ thermal stages. All stages under precursor candidate $p_k$ use
$(\hat T_k,\hat t_k,\hat a_k)$, so a route run alone occupies
$H_r\hat t_k$ dwell-hours. Atmosphere has two compatibility classes, \texttt{air} and \texttt{nonair}.
Missing atmosphere defaults to \texttt{air} for scheduling but remains
omitted from SOP prompts; atmosphere is excluded from retrieval weighting.
Non-furnace operations retain their sequence order but do not constrain
furnace availability in the simulation.

\paragraph{Compatibility and cost.}
Thermal stages execute in rounds: all first stages finish before any
second stages begin, and similarly for later stages. Only stages with the
same index $h$, numerical time $t$, and atmosphere $a$ can share a run.
Within each block $b=(h,t,a)$, a batch $A$ contains at most $C$ samples and
requires a common intersection of the intervals
$[\hat T_r-\Delta_F,\hat T_r+\Delta_F]$:
\begin{equation}
\max_{r\in A}\hat T_r-\min_{r\in A}\hat T_r\leq2\Delta_F.
\label{eq:temperature_batch_intersection}
\end{equation}
The executed setpoint is the intersection midpoint,
$\theta_A=(\min_{r\in A}\hat T_r+\max_{r\in A}\hat T_r)/2$.
A chain of overlapping neighboring intervals does not ensure a common intersection.
Numerical times must agree exactly; SOP prompt intervals and time bins
are not used for batching.

Let $n_b^*(S;\Delta_F)$ denote the minimum number of compatible
capacity-$C$ batches for selected routes $S$ in block $b$. Then
\begin{equation}
\begin{aligned}
C_{\mathrm{run}}(S;\Delta_F)&=\sum_b n_b^*(S;\Delta_F),\\
C_{\mathrm{hour}}(S;\Delta_F)&=\sum_b (\alpha+t_b)n_b^*(S;\Delta_F).
\end{aligned}
\label{eq:interval_furnace_costs}
\end{equation}
For a common $\Delta_F$, sorting temperatures and repeatedly taking the
largest prefix within capacity and temperature span minimizes the run
count. An exchange argument puts the earliest stages in the first batch
without increasing the number of batches and applies recursively.
Since each run in block $b$ costs $\alpha+t_b\geq0$, the same partition
minimizes furnace-hours. This rule reconstructs both heuristic and MILP
plans, preserving run membership for scoring. A route with no thermal
stages consumes one trial and zero furnace resources.

The main simulation uses $\Delta_F=0$, $C=8$, and per-run overhead
$\alpha=2$ h. Overhead is a benchmark assumption, not a measured equipment
constant.
Positive temperature tolerances are evaluated in
Appendix~\ref{app:temperature_milp}. Furnace-hours measure occupancy under
these programs, excluding temperature-dependent ramp costs and other
laboratory time; parallel furnace count is not constrained.

\paragraph{Invariance to stage programs.}
Run counts depend on predicted conditions only through which stages at
the same position can share a run. At $\Delta_F=0$, sharing requires
identical temperature, time, and atmosphere, so any stage program that sets
the stage-$h$ temperature and time to strictly increasing functions
$f_h(\hat T_k)$ and $g_h(\hat t_k)$, such as a lower first-stage calcination
temperature or a shorter first-stage dwell, preserves compatibility. At
$\Delta_F=50\,{}^\circ$C, the same holds when $f_h$ shifts temperatures by a
stage-specific constant, which preserves temperature spans within each
stage position. In both cases every run-budget plan and its recovery are
unchanged; only furnace-hours can change. Programs that depend on the
operation label or on a route's total stage count are not covered by this
argument.

\paragraph{Heuristic selectors.}
Order by PP rank sorts by precursor rank and then SOP rank; Order by
SOP rank sorts by SOP rank and then precursor rank. Both use the full
\proposed{} candidate pool, whereas MSP-LLM selection restricts admission
to the Top-1 precursor branch. Utility-greedy sorts by
Eq.~\eqref{eq:route_utility}.
Each accepts a route only if all its stages fit every active budget after
repacking; an infeasible candidate is skipped and the scan continues.
A bundle heuristic supplies additional feasible incumbents by considering utility-ordered prefixes with identical program/stage-count profiles and scoring marginal utility against marginal resource use. Overlapping temperature intervals alone do not define a
bundle profile.

\paragraph{MILP selection and allocation.}
Let $x_r\in\{0,1\}$ select route $r$. In block $b$, group stages with the
same temperature into classes $c$, with $\mathcal R_c$ the routes
contributing a stage to class $c$. Let $\mathcal E_b$ contain the maximal
contiguous neighborhoods of classes spanning at most $2\Delta_F$, and
let $\mathcal E=\bigcup_b\mathcal E_b$ retain block identities.
For $c\in e$, $z_{ce}\geq0$ assigns class demand to neighborhood $e$,
$y_e\in\mathbb Z_{\geq0}$ counts its runs, and $t_e$ is their common time:
\begin{equation}
\begin{aligned}
\max_{x,z,y}\quad &\sum_{r\in\mathcal R_m}u(r)x_r\\
\mathrm{s.t.}\quad
&\sum_{e:c\in e}z_{ce}=\sum_{r\in\mathcal R_c}x_r
&&\forall c,\\
&\sum_{c\in e}z_{ce}\leq C y_e &&\forall e,\\
&\sum_r x_r\leq B_{\mathrm{trial}} &&\text{if }\mathrm{trial}\in\mathcal Q,\\
&\sum_e y_e\leq B_{\mathrm{run}} &&\text{if }\mathrm{run}\in\mathcal Q,\\
&\sum_e(\alpha+t_e)y_e\leq B_{\mathrm{hour}} &&\text{if }\mathrm{hour}\in\mathcal Q,\\
&x_r\in\{0,1\},\quad y_e\in\mathbb Z_{\geq0},\quad z_{ce}\geq0.
\end{aligned}
\label{eq:route_batch_milp}
\end{equation}
Selecting a route creates demand at every thermal stage, prohibiting
partial execution. Every feasible batch lies in a maximal neighborhood,
and any group of at most $C$ stages in one neighborhood admits a common
setpoint. Continuous $z$ preserves feasibility: for fixed integer $x,y$,
the assignment is a bipartite flow with integer demands and capacities,
so a feasible fractional flow has an integral counterpart.
Single-neighborhood blocks eliminate assignment variables. Identical
resource-demand profiles permit utility-sorted prefix restrictions;
these are not applied to profiles that merely overlap in temperature.
Selected sets are repacked by Eq.~\eqref{eq:interval_furnace_costs}, giving
all selectors the same cost and executed-setpoint convention.

\paragraph{Solver limits and certification.}
SciPy/HiGHS\citep{huangfu2018parallelizing} uses one CPU thread, a 2 s solver limit per call, relative gap
tolerance $10^{-9}$, feasibility tolerance $10^{-8}$, a fixed solver seed,
and presolve disabled. Utilities are rescaled by a common positive factor.
We retain the best feasible incumbent from the solver and utility/bundle
heuristics, checking all stages, compatibility, capacity, and active
budgets. Certification compares the reconstructed incumbent with the
solver bound within numerical tolerances; a timeout retains its feasible
plan and bound without an optimality claim unless the bound is matched.
On the original benchmark, all main repeated-program and matched-pool
cases are certified.

\paragraph{Computational cost.}
\label{app:planning_runtime}
Database conditions and the precursor-to-procedure lookup are precomputed.
Table~\ref{tab:app_planning_runtime} reports optimization times for
$\Delta_F=0$, $C=8$, $\alpha=2$ h, pooling 918 targets and three seeds
without reused compatibility solutions. Timing includes utility/bundle
incumbent construction and MILP formulation and solution on one solver
thread. These shared-host wall-clock measurements exclude PP, SCP, SOP,
and retrieval, and therefore do not measure end-to-end latency.
\begin{table}[!htbp]
\centering
\begingroup\color{black}\small
\captionsetup{font={color=black},labelfont={color=black}}
\caption{Planning optimization runtime.}
\label{tab:app_planning_runtime}
\begin{tabular}{lrrr}
\toprule
Budget & Target--seed cases & Median (s) & 95th percentile (s) \\
\midrule
3 runs & 2,754 & 0.0109 & 0.0292 \\
20 runs & 2,754 & 0.0464 & 0.1337 \\
48 h & 2,754 & 0.0155 & 0.0597 \\
\bottomrule
\end{tabular}
\endgroup
\end{table}

\paragraph{Reference scoring.}
References are used only for scoring, never to select routes or change
execution programs and costs. Recipe recovery requires a selected route
to match the precursor set and ordered operations of the same reference.
\endgroup

\endgroup

\FloatBarrier
\begingroup
\section{Dataset}
\label{app:dataset}

\subsection{Original benchmark and condition annotations}
\label{app:original_dataset}
The original split comprises 8,295 training, 932 validation, and 996 test
recipes. Inference uses all test recipes; evaluation uses 918 fixed
representatives with one reference recipe per target.
The data use seven operation labels:
heating, sintering, annealing, mixing, shaping, quenching, and drying.
The training split contains 285 distinct precursors.
Article-title synthesis context is excluded from every model input.

The 27,110-record condition-training corpus comprises 6,664 records from
the SOP training split and 20,446 additional condition-labeled records.
The entire condition-training corpus, including these additional records,
excludes targets in the original validation and test splits, using
the original benchmark's target identifiers.
Table~\ref{tab:app_condition_annotation_rates} reports field availability,
with overlapping counts for individual fields and a separate count for
complete triples. Training responses omit missing fields rather than requiring complete
triples. Temperature and time targets are the maximum recorded values,
and atmosphere is evaluated as air/nonair. Missing annotations do not
remove targets from SOP or full-planning evaluation.

\begin{table}[!htbp]
\centering
\begingroup\color{black}\small
\captionsetup{font={color=black},labelfont={color=black}}
\caption{Synthesis-condition annotation availability.}
\label{tab:app_condition_annotation_rates}
\begin{tabular}{lrr}
\toprule
Observed annotations & Records & Share (\%) \\
\midrule
Temperature & 25,963 & 95.77 \\
Time & 24,023 & 88.61 \\
Atmosphere & 13,058 & 48.17 \\
All three conditions & 11,277 & 41.60 \\
\bottomrule
\end{tabular}
\endgroup
\end{table}

\subsection{New target set: source and filtering}
\label{app:unseen_dataset}
\label{app:unseen_construction}
We construct a separate evaluation dataset from the 80,806 solid-state
synthesis records released by \citet{lee2025text}. We remove targets overlapping the original
splits or condition-training corpus, symbolic or unparseable formulas,
ambiguous or out-of-vocabulary precursors, elemental targets, records
without synthesis actions, and duplicate records.
Target identity is the vector of elemental atomic fractions rounded to ten
decimals. A conservative element-support filter also excludes potential
overlaps hidden in symbolic training formulas.
Precursor spellings are normalized to the training vocabulary only when
absolute atom counts agree; hydration and stoichiometry are preserved.

\subsection{Operation curation and verification}
\label{app:unseen_curation}
\label{app:unseen_verification}
LLM-assisted curation (\texttt{gpt-5.4-mini-2026-03-17}, medium reasoning)
separates synthesis steps from characterization, repeated descriptions,
and actions for other samples. Inputs contain extracted event fragments
for the specified target--precursor pair. Each step cites source events,
and every event is assigned to a step or an explicit drop reason.
Thermal labels follow the recorded action token: sinter maps to sintering,
anneal to annealing, and other thermal actions to heating.

Two independent responses must agree on the ordered sequence after collapsing
adjacent identical labels. A third source-fidelity review approves, flags,
or rejects each record without rewriting it. We reject invented or reordered
steps, merges across separately recorded stages, and sequences without a
thermal operation, without imposing a preferred recipe length.
The resulting dataset contains 2,482 distinct paired references for
2,344 targets: 2,236 single-reference targets and 108 with multiple references.

\paragraph{Paired-reference scoring.}
Each target contributes one recovery outcome. A selected route
must match both the precursor set and the ordered operations of the same
reference; a match to any one of the target's paired references satisfies this criterion.
Precursor matches and operation matches from different references cannot be
combined. For this new-set evaluation, adjacent repeated operation labels are
collapsed in both predictions and references. Targets with invalid or empty
outputs remain in the denominator. Models and fusion calibrators are frozen
before evaluation on the new targets.

Programmatic checks verify target disjointness, precursor-vocabulary membership,
unique reference pairs, thermal-operation presence, and event provenance.
Recorded condition fragments are retained as evidence but are not used
as ground truth in the new-set furnace evaluation.

\subsection{Dataset characteristics}
\label{app:unseen_characteristics}
Table~\ref{tab:app_unseen_profile} compares reference-recipe characteristics,
using the 918 evaluated representatives for the original test column.
The new set contains more non-oxide targets and all-elemental precursor
sets, and shorter operation sequences, testing generalization under a
chemistry and procedure shift. Here, ``oxide'' denotes an oxygen-containing
target formula; each reference recipe contributes to the profile statistics.
We compare the frozen methods on the same new targets under common
paired-reference scoring.
The split-specific chemistry results are in
Appendix~\ref{app:new_target_results}.
\endgroup
\begin{table}[!htbp]
\centering
\caption{\textbf{Recipe characteristics of the original and new data.}}
\label{tab:app_unseen_profile}
\begingroup\footnotesize\setlength{\tabcolsep}{3pt}\papertablesetup
\begin{tabular*}{\linewidth}{@{\extracolsep{\fill}}lrrrrr@{}}
\toprule
 & \makecell{Original\\train} & \makecell{Original\\test (918)} & \makecell{New\\all refs.} & \makecell{New\\single refs.} & \makecell{New\\multi refs.} \\
\midrule
Recipes & 8,295 & 918 & 2,482 & 2,236 & 246 \\
Oxide (\%) & 92.5 & 92.4 & 32.6 & 33.9 & 21.1 \\
All-elemental precursors (\%) & 6.6 & 6.3 & 64.6 & 63.2 & 77.6 \\
Mean operation count & 4.17 & 4.31 & 2.71 & 2.72 & 2.66 \\
One-operation recipes (\%) & 3.1 & 1.6 & 23.5 & 23.4 & 24.4 \\
Mean precursor count & 3.23 & 3.31 & 3.56 & 3.58 & 3.39 \\
Mean elements per target & 4.18 & 4.25 & 3.89 & 3.93 & 3.61 \\
\bottomrule
\end{tabular*}\endgroup
\end{table}

\FloatBarrier

\FloatBarrier
\begingroup
\section{Baselines}
\label{app:baselines}
\label{app:table_notes}

\subsection{Condition prediction}
The predictor of \citet{huo2022machine} is adapted to temperature and time regression;
SyntMTE \citep{prein2025language} is adapted to temperature prediction.
Both are retrained with their own implementations.
The predictor of \citet{huo2022machine} uses the original training split
augmented with parseable condition records from the source corpus;
SyntMTE uses the original training split.
Table~\ref{tab:app_external_data} additionally reports SyntMTE with
LLM-generated training data and a GBM using all available temperature labels.
Unsupported tasks are marked N/A.
The GBM baseline uses the same composition experts as \proposed, without
the joint LLM.

\subsection{Operation prediction and full planning}
The MPC baseline adapts the learned material-similarity retriever of
\citet{he2023precursor} to return operation sequences from retrieved procedures.
SPENDE \citep{na2023artificial} returns one operation sequence, so its
Top-$k$ results for $k>1$ are unavailable.
MSP-LLM uses the same backbone and SOP training recipe as our generator
but receives no condition block.
Standalone SOP uses the reference precursors for every method.
In full planning, the original MSP-LLM strategy uses the highest-ranked
precursor set from the shared PP model and admits its SOP candidates in order.
For MSP-LLM with \proposed{} selection in
Tables~\ref{tab:main_msp_budgets} and~\ref{tab:main_unseen_trials}, the same
frozen SOP model generates 30 beams for each of the Top-10 predicted
precursor sets, using its original prompt and decoding settings without
condition context or retrieval fusion. The resulting alternatives receive
the utility-based selection of \proposed{}.
Conversely, \proposed{} with MSP-LLM selection admits its fused SOP
candidates in order within the Top-1 precursor branch.
The Order by PP rank and Order by SOP rank controls in
Table~\ref{tab:main_ablation}(c) both use the full \proposed{} pool.

\subsection{API-based LLMs}
GPT-4o mini uses \texttt{gpt-4o-mini-2024-07-18}, fine-tuned on the same
8,295 training recipes. Separate fine-tunes predict precursors, operations,
or the recipe. SOP and recipe queries draw 30 samples
at temperature 0.8, ordered by total sequence log-probability.
The two-stage baseline passes the greedy precursor prediction to the SOP
fine-tune. Its error bars summarize API request seeds 42, 46, and 51,
not independent fine-tuning runs.

GPT-5.4 mini uses zero-shot or 100-shot prompting without fine-tuning;
demonstrations are fixed training examples sampled once with seed 0.
SOP and recipe queries request 30 ordered candidates in one response at
temperature 1.0 with reasoning disabled. End-to-end outputs contain precursors and operations; the
two-stage variant first predicts one precursor set at temperature 0.
We report request seeds 42 and 46, limiting this baseline to two repetitions because of API cost.

\paragraph{Candidate handling.}
Duplicate or malformed raw outputs consume slots in trial-budget and
standalone SOP evaluations. Before furnace scheduling, malformed routes
are excluded and valid duplicates are retained for the baselines.
In full planning, all systems use predicted precursors and the full target denominator,
including incorrect top-1 precursor predictions. They share the program
predictor and furnace cost model at each $\Delta_F$.
The original MSP-LLM and API baselines admit whole routes in candidate
order. Rows using \proposed{} selection use the same utility-based MILP
procedure; rows using MSP-LLM selection retain Top-1-branch admission.
Main furnace results use $\Delta_F=0$.
Utility-greedy and relaxed-sharing comparisons appear in
Appendix~\ref{app:temperature_milp}, with common fixed-set batch costs at
each tolerance. Since system candidate pools differ, the same-pool controls
in Table~\ref{tab:main_ablation} isolate the selection rule.

\subsection{Precursor prediction (PP)}
\label{app:shared_pp}
Both systems share the frozen MSP-LLM precursor model
(training seed 42; Appendix~\ref{app:training_details}).
Table~\ref{tab:app_shared_pp} reports precursor-set exact match (\%) for its
ten-beam outputs on the original test and new targets.
Set matching ignores order and repeated formulas, strips whitespace and
case, and does not substitute chemically related compounds. Invalid
predictions count as misses.

\begin{table}[!htbp]
\centering
\begingroup
\footnotesize\setlength{\tabcolsep}{4pt}\papertablesetup
\caption{\textbf{Shared precursor prediction.}}
\label{tab:app_shared_pp}
\resizebox{\linewidth}{!}{%
\begin{tabular}{llrrrrr}
\toprule
Dataset (targets) & Ranking & Top-1 & Top-2 & Top-3 & Top-5 & Top-10 \\
\midrule
Original test (918) & Original beams & 74.29 & 84.10 & 88.34 & 91.83 & 94.55 \\
 & Distinct sets & 74.29 & 84.42 & 88.78 & 92.16 & 94.55 \\
\midrule
New targets (2,344) & Original beams & 83.19 & 87.41 & 89.93 & 91.68 & 93.90 \\
 & Distinct sets & 83.19 & 87.93 & 90.32 & 91.98 & 93.90 \\
\bottomrule
\end{tabular}}
\par\vspace{3pt}\begin{minipage}{\linewidth}\footnotesize
Both systems share one frozen PP checkpoint; no between-SOP-seed SD applies.
Each target is counted once. The original test uses one reference per target;
new targets match any precursor set among their 2,482 reference
recipes (2,236 single-reference and 108 multi-reference targets).
Original beams retain duplicate slots; distinct sets merge repeats in
first-appearance order within the same ten outputs. Invalid candidates
count as misses; no additional beams are generated.
\end{minipage}
\endgroup
\end{table}

On the original set, top-1 is correct for 682 targets and top-10 contains
the reference precursor set for 868. Thus, 186 of the 236 top-1 errors have
a correct alternative in the pool, while 50 targets remain uncovered.
New-set top-1 and top-10 are 83.19\% and 93.90\%, respectively.

\subsection{Adaptation of the interpretable condition predictor}
\label{app:karpovich_adaptation}
We adapt the chemistry-only \texttt{mp\_fraction} XGBoost regressors of
\citet{karpovich2023interpretable}. Features are target elemental fractions,
precursor functional-group fractions, and precursor melting-point
minimum, maximum, mean, and range; missing melting points use the released
lookup's median. Operation sequences and condition labels are excluded
from inputs. Separate regressors predict maximum temperature and time in
native units; the implementation has no atmosphere output.

Training combines the original split with raw condition records, as for the predictor of \citet{huo2022machine}, excluding unparseable formulas and validation/test targets by exact or
reduced formula. Following the official aggregation, labels for identical
target--precursor sets are averaged, yielding 7,670 temperature and 7,333
time groups. The respective 712 and 717 validation labels are held out.
We select configurations by validation RMSE using 100 Hyperopt TPE trials
per variable and seed (42, 46) over the official search space:
\begin{center}
\small
\begin{tabular}{ll}
\toprule
Parameter & Candidate values \\
\midrule
Trees & 25, 50, 100, 200 \\
Maximum depth & 3, 6, 9 \\
Learning rate & 0.1, 0.2, 0.3 \\
Column fraction (tree and level, independently) & 0.6, 0.8, 1 \\
Minimum child weight & 1, 2, 3 \\
Row fraction & 0.7, 0.9, 1 \\
\bottomrule
\end{tabular}
\end{center}
All 996 test inputs receive predictions; evaluation follows
Appendix~\ref{app:evaluation_details}.
\endgroup

\FloatBarrier
\section{Additional Experiments}
\label{app:additional_experiments}

\subsection{Training data for condition baselines}
\label{app:scp_details}
Table~\ref{tab:app_external_data} reports the condition-baseline training
configurations. All condition-training pools exclude targets in the original
validation and test splits, including the 20,446 additional records in
our condition corpus. Thus, condition training introduces no validation/test
target leakage under the original benchmark's target identifiers.
Raw-corpus augmentation gives the adapted baselines access to the same
27,110-record source as the joint model, retaining the required labels
and formulas supported by each implementation. Smaller usable sets reflect
these annotation and preprocessing requirements. The joint model instead
uses available fields without requiring every field to be annotated.

We additionally report SyntMTE trained on the authors' literature and
LLM-generated LLMForge recipes, and on their combination with our augmented
training data. These are pooled-data adaptations to maximum-temperature
prediction: both temperature heads regress the record-level maximum and
their outputs are averaged.
The combined-data variant uses our validation split; the authors'-data
variant retains its source validation split.

The full-label GBM uses all 25,963 temperature-annotated records from our
condition corpus, with the same composition descriptors, absolute-error
loss, learning rate 0.06, and 300 boosting iterations as the original
expert. Unavailable descriptor blocks are represented as missing values,
which the histogram-based GBM handles natively, so no labeled record is
dropped solely because its formula cannot be featurized. Original formula
strings and temperature labels are unchanged. All rows are evaluated using
the common temperature protocol of Table~\ref{tab:main_scp}.
Stars mark the configurations used in that table.
\begin{table}[!htbp]
\centering
\caption{Training data for adapted condition baselines.}
\label{tab:app_external_data}
\begingroup\footnotesize\setlength{\tabcolsep}{3pt}\papertablesetup
\begin{tabular*}{\linewidth}{@{\extracolsep{\fill}}llrrr@{}}
\toprule\rowcolor{gray!12}\multicolumn{5}{l}{\textit{Temperature}} \\
Model & Training data & Rows & MAE $\downarrow$ & RMSE $\downarrow$ \\
\midrule
\citet{huo2022machine} & Original split & 6,228 & 152.84 & 197.16 \\
\citet{huo2022machine} & + raw corpus $\star$ & 11,331 & 151.76 & 195.29 \\
\midrule
SyntMTE & Original split $\star$ & 6,228 & \vsd{171.79}{8.71} & \vsd{217.60}{11.65} \\
SyntMTE & + raw corpus & 11,331 & \vsd{173.07}{2.83} & \vsd{219.20}{2.53} \\
SyntMTE & Literature + LLM data & 30,762 & \vsd{180.85}{5.72} & \vsd{228.02}{10.78} \\
SyntMTE & Above + our corpus & 38,390 & \vsd{178.76}{2.22} & \vsd{221.72}{0.36} \\
\midrule
GBM & Original split & 6,228 & 143.42 & 190.88 \\
GBM & All temperature labels & 25,963 & 144.48 & 192.11 \\
\midrule
\proposed{} (threshold) & Joint / expert & 27,110 / 6,228 & \vsd{\textbf{131.77}}{3.15} & \vsd{\textbf{182.03}}{0.87} \\
\bottomrule
\end{tabular*}
\par\vspace{4pt}
\begin{tabular*}{\linewidth}{@{\extracolsep{\fill}}llrrrr@{}}
\toprule\rowcolor{gray!12}\multicolumn{6}{l}{\textit{Time}} \\
Model & Training data & Rows & MAE $\downarrow$ & RMSE $\downarrow$ & $\pm3$h $\uparrow$ \\\midrule
\citet{huo2022machine} & Original split & 6,255 & 13.13 & 25.73 & 32.38 \\
\citet{huo2022machine} & + raw corpus $\star$ & 10,627 & 13.11 & 25.64 & 32.06 \\
\proposed{} (threshold) & Joint / expert & 27,110 / 6,255 & \vsd{\textbf{12.43}}{0.35} & \vsd{\textbf{24.41}}{0.69} & \vsd{\textbf{40.08}}{1.46} \\
\bottomrule\end{tabular*}\endgroup
\par\vspace{3pt}\begin{minipage}{\linewidth}\footnotesize
All training pools exclude original validation/test targets.
Raw-corpus augmentation uses the joint model's 27,110-record source;
usable counts depend on required labels and formula preprocessing.
LLM data denotes the authors' synthetic LLMForge recipes; ``Above + our
corpus'' combines their literature and synthetic data with our augmented
training set. The full-label GBM retains all 25,963 temperature labels,
using native missing-feature handling for unavailable descriptors.
Stars identify the configurations in Table~\ref{tab:main_scp}.
\proposed{} lists joint-model and expert training sizes separately.
\end{minipage}
\end{table}

\subsection{Soft operation-sequence metrics}
\label{app:soft_metrics}
Table~\ref{tab:app_sop_soft} uses the same candidates as
Table~\ref{tab:main_sop}, with reference precursors on all 918 targets.
Scores are percentages, with higher values indicating closer agreement.
For a prediction $o$ and reference $o^\star$,
NED similarity is $1-d_{\rm edit}(o,o^\star)/\max(|o|,|o^\star|)$;
LCS-F1 is $2\,{\rm LCS}(o,o^\star)/(|o|+|o^\star|)$; multiset-F1
replaces LCS with the multiset intersection. Each Top-$k$ score is the
best value among the first $k$ candidates. These metrics give partial
sequence credit without assessing chemical validity.
\begin{table}[!htbp]
\centering
\caption{\textbf{Soft operation-sequence scores.}}
\label{tab:app_sop_soft}
\begingroup\footnotesize\setlength{\tabcolsep}{3pt}\papertablesetup
\resizebox{\linewidth}{!}{%
\begin{tabular}{lrrrrrr}
\toprule
 & \multicolumn{2}{c}{NED similarity} & \multicolumn{2}{c}{LCS-F1} & \multicolumn{2}{c}{Multiset-F1} \\
\cmidrule(lr){2-3}\cmidrule(lr){4-5}\cmidrule(lr){6-7}
Method & Top-10 & Top-30 & Top-10 & Top-30 & Top-10 & Top-30 \\
\midrule\rowcolor{gray!12}\multicolumn{7}{l}{\textit{Specialized models}} \\
MPC & \msd{69.51}{0.10} & \msd{78.87}{0.17} & \msd{79.38}{0.12} & \msd{86.37}{0.15} & \msd{83.43}{0.26} & \msd{90.10}{0.26} \\
SPENDE & N/A & N/A & N/A & N/A & N/A & N/A \\
MSP-LLM & \msd{75.08}{0.12} & \msd{83.67}{0.02} & \msd{83.22}{0.02} & \msd{89.27}{0.06} & \msd{86.81}{0.15} & \msd{92.26}{0.07} \\
\midrule\rowcolor{gray!12}\multicolumn{7}{l}{\textit{Large language models}} \\
GPT-5.4 mini (zero-shot) & \msd{53.19}{0.43} & \msd{62.86}{0.15} & \msd{65.68}{0.38} & \msd{74.81}{0.05} & \msd{71.18}{0.19} & \msd{81.13}{0.14} \\
GPT-5.4 mini (100-shot) & \msd{67.22}{0.15} & \msd{75.79}{0.34} & \msd{78.08}{0.19} & \msd{84.58}{0.30} & \msd{82.04}{0.03} & \msd{88.45}{0.20} \\
GPT-4o mini (fine-tuned) & \msd{62.48}{0.15} & \msd{77.31}{0.09} & \msd{73.71}{0.16} & \msd{85.17}{0.16} & \msd{76.53}{0.23} & \msd{88.83}{0.25} \\
\midrule
\proposed{} & \msd{76.51}{0.29} & \msd{84.76}{0.11} & \msd{84.35}{0.13} & \msd{90.13}{0.09} & \msd{87.52}{0.07} & \msd{92.97}{0.09} \\
\bottomrule\end{tabular}}\endgroup
\par\vspace{3pt}\begin{minipage}{\linewidth}\footnotesize
SD uses three training seeds for MPC, SPENDE, MSP-LLM, and Ours; three
request seeds for GPT-4o mini and two for GPT-5.4 mini.
SPENDE returns one sequence: its Top-1 NED, LCS-F1, and multiset-F1 are
$50.48\pm0.75$, $63.68\pm0.60$, and $65.87\pm0.84$.
Its higher-$k$ scores are unavailable, as in Table~\ref{tab:main_sop}.
\end{minipage}
\end{table}

\subsection{New-target results by chemistry}
\label{app:new_target_results}
Table~\ref{tab:app_unseen_chem_split} reports paired-reference recipe
recovery (\%) at 1, 10, and 30 trials on oxygen-containing (oxide, $n=781$)
and non-oxygen-containing targets ($n=1,563$). The elemental-precursor subset
($n=1,495$) requires every reference to use only elemental precursors and
overlaps these groups. Values are mean $\pm$ sample SD across SOP training
seeds, or API request seeds for GPT models.
At 30 trials, \proposed{} exceeds MSP-LLM by 4.44 points on
oxide targets and 7.97 points on non-oxide targets, showing that the
gain extends beyond the dominant training chemistry.
\providecommand{\vsd}[2]{\begin{tabular}[t]{@{}r@{}}#1\\[-2pt]{\scriptsize($\pm$\,#2)}\end{tabular}}
\begin{table}[!htbp]
\centering
\caption{\textbf{New-target recipe recovery by chemistry.}}
\label{tab:app_unseen_chem_split}
\begingroup\footnotesize\setlength{\tabcolsep}{2.5pt}\papertablesetup\renewcommand{\arraystretch}{1.0}
\resizebox{\linewidth}{!}{%
\begin{tabular}{l rrr rrr rrr}
\toprule
 & \multicolumn{3}{c}{Oxide} & \multicolumn{3}{c}{Non-oxide} & \multicolumn{3}{c}{Elemental precursors} \\
\cmidrule(lr){2-4}\cmidrule(lr){5-7}\cmidrule(lr){8-10}
Method & 1 & 10 & 30 & 1 & 10 & 30 & 1 & 10 & 30 \\
\midrule
MSP-LLM & \vsd{\textbf{6.23}}{0.45} & \vsd{20.10}{0.56} & \vsd{30.30}{0.64} & \vsd{3.71}{0.55} & \vsd{8.81}{0.18} & \vsd{17.47}{1.33} & \vsd{3.61}{0.57} & \vsd{8.72}{0.22} & \vsd{17.59}{1.34} \\
\addlinespace[2pt]
GPT-4o mini (end-to-end, fine-tuned) & \vsd{5.93}{0.07} & \vsd{13.83}{0.13} & \vsd{21.25}{1.79} & \vsd{3.86}{0.07} & \vsd{5.25}{0.28} & \vsd{17.27}{1.56} & \vsd{3.97}{0.10} & \vsd{5.35}{0.29} & \vsd{17.81}{1.59} \\
\addlinespace[2pt]
GPT-4o mini (PP$\rightarrow$SOP, fine-tuned) & \vsd{5.68}{0.32} & \vsd{10.84}{0.58} & \vsd{18.57}{0.13} & \vsd{3.13}{0.06} & \vsd{4.61}{0.17} & \vsd{16.59}{0.87} & \vsd{3.14}{0.07} & \vsd{4.59}{0.17} & \vsd{16.88}{0.91} \\
\addlinespace[2pt]
GPT-5.4 mini (end-to-end, zero-shot) & \vsd{0.06}{0.09} & \vsd{1.28}{0.18} & \vsd{4.74}{0.18} & \vsd{0.58}{0.18} & \vsd{2.21}{0.14} & \vsd{4.29}{0.09} & \vsd{0.60}{0.19} & \vsd{2.27}{0.09} & \vsd{4.41}{0.09} \\
\addlinespace[2pt]
GPT-5.4 mini (end-to-end, 100-shot) & \vsd{5.19}{0.45} & \vsd{13.25}{0.45} & \vsd{18.44}{0.36} & \vsd{4.13}{0.23} & \vsd{6.37}{0.14} & \vsd{8.22}{0.14} & \vsd{4.21}{0.28} & \vsd{6.09}{0.09} & \vsd{7.83}{0.28} \\
\addlinespace[2pt]
GPT-5.4 mini (PP$\rightarrow$SOP, zero-shot) & \vsd{0.00}{0.00} & \vsd{2.11}{0.63} & \vsd{8.64}{0.09} & \vsd{0.16}{0.05} & \vsd{3.71}{0.18} & \vsd{9.47}{0.54} & \vsd{0.17}{0.05} & \vsd{3.85}{0.24} & \vsd{9.57}{0.57} \\
\addlinespace[2pt]
GPT-5.4 mini (PP$\rightarrow$SOP, 100-shot) & \vsd{4.93}{0.09} & \vsd{11.97}{0.09} & \vsd{17.35}{0.09} & \vsd{\textbf{4.57}}{0.05} & \vsd{9.98}{0.45} & \vsd{14.94}{0.59} & \vsd{\textbf{4.58}}{0.14} & \vsd{10.00}{0.33} & \vsd{15.18}{0.47} \\
\addlinespace[2pt]
\midrule
\proposed{} & \vsd{6.10}{0.20} & \vsd{\textbf{21.25}}{0.26} & \vsd{\textbf{34.74}}{0.52} & \vsd{4.03}{0.61} & \vsd{\textbf{11.60}}{0.10} & \vsd{\textbf{25.44}}{0.33} & \vsd{3.97}{0.61} & \vsd{\textbf{11.37}}{0.07} & \vsd{\textbf{25.64}}{0.38} \\
\bottomrule
\end{tabular}}
\endgroup
\end{table}

\subsection{Utility and confidence-dependent allocation}
\label{app:utility_rationale}
PP mass compares precursor support across branches, while reciprocal SOP
rank applies the same gradual depth discount within each branch. The rank
already reflects fused generation and retrieval evidence, without requiring
fusion-score gaps to represent comparable gains across precursor branches.

\paragraph{Direct fusion-score utility.}
\label{app:fusion_score_utility}
Replacing $\pi_k^{\mathrm{PP}}/j$ with
$\pi_k^{\mathrm{PP}}s_{\mathrm{fuse}}(o_{kj})$ while fixing the candidate
pool, MILP, and costs gives similar recovery on the 918-target test set:
40.01\% versus 39.87\% at 30 trials and 31.99\% versus 32.24\% at 48 h
(reciprocal rank versus direct scores; means over three SOP seeds).
All compared plans are certified for their utilities. Reciprocal rank
provides a simple common depth discount without requiring comparable
fusion-score gaps across precursor sets.

\subsection{Selection algorithms and sharing tolerance}
\label{app:temperature_milp}

\paragraph{Utility-greedy and MILP selection.}
We fix the \proposed{} candidate pool, condition predictions,
PP masses, SOP ranks, capacity $C=8$, and overhead $\alpha=2$ h.
Utility-greedy admits whole routes in utility order whenever their stages
fit the budget; MILP optimizes the same utility under shared execution
costs. MILP was selected by validation recovery, weighting the run- and
hour-budget families equally at each tolerance.
At $\Delta_F=0$, greedy and MILP recover 30.25\% and 32.90\% at three runs,
and 30.79\% and 31.99\% at 48 h, respectively, averaged across three SOP
seeds on 918 targets. At one run, greedy is slightly higher
(21.86\% versus 21.71\%). Under trial budgets alone, both use the same
utility ranking and recover 40.01\% at 30 trials.
Both selectors thus optimize the same route utility and coincide under trial budgets; under furnace budgets, MILP generally adds recovery by accounting for shared runs.
Tables~\ref{tab:main_msp_budgets} and~\ref{tab:main_unseen_trials}
cross candidate construction with selection, while
Table~\ref{tab:main_ablation}(c) compares selection strategies on the same
full candidate pool.

\paragraph{Relaxed temperature sharing.}
Keeping the same pool, utilities, and exact time, atmosphere, and
stage-position matching, $\Delta_F=50\,{}^\circ$C gives greedy/MILP
recovery of 31.19/33.66\% at three runs and 31.45/32.46\% at 48 h,
compared with 30.25/32.90\% and 30.79/31.99\% at $\Delta_F=0$;
trial-only selection is unchanged. Two observations follow. First,
the MILP advantage over greedy persists at the looser tolerance
(2.47 points at three runs; 1.02 points at 48 h, 95\% CI
[$-0.18$, 2.18], where the interval includes zero), indicating that
the gain from joint selection does not depend on exact-temperature
matching. Second, relaxing the tolerance raises recovery for both
selectors by 0.5--0.9 points, because more stages become batchable;
the effect is modest because conditions are shared within each precursor
branch and sharing across branches also requires identical predicted times. A larger tolerance does not guarantee higher
recovery: the shared setpoint deviates from each route's predicted
temperature by up to $\Delta_F$, and the selected routes can change, so
a plan that is cheaper under the cost model need not contain more
reference routes. We therefore report $\Delta_F=0$ as the main setting
and treat $\Delta_F$ as a scenario parameter whose appropriate value
depends on how much setpoint deviation the target chemistry tolerates.
Confidence intervals use 2,000 target-paired bootstrap draws after
averaging the three seed outcomes within each target; they are
pointwise.

\subsection{Budget-wise admission}
\label{app:furnace_details}
\label{app:furnace_nonmonotone}
Each budget is evaluated independently. Ordered-admission baselines
scan their candidate lists from the beginning, admitting a whole route
when all its stages fit the current budget. MILP solves the corresponding
utility-allocation problem at each budget. We score the returned plan
for that budget without substituting a plan evaluated at another budget.

\FloatBarrier
\subsection{Paired uncertainty in SOP comparisons}
\label{app:sop_paired_uncertainty}
Table~\ref{tab:app_sop_paired_ci} reports paired differences for the SOP
comparisons in Tables~\ref{tab:main_sop}.
All methods are evaluated on the same 918 test targets with reference
precursors. For each target, we first average the exact-match difference
over the three matched SOP seeds, then resample the 918 targets with
replacement 20,000 times. We report percentile 95\% confidence intervals.
These pointwise intervals quantify target sampling uncertainty conditional
on the fitted models; training variation is reported separately as seed
SD in the main tables. No multiple-comparison adjustment is applied.

Relative to \proposed{} without conditions, the gains at Top-10, Top-20,
and Top-30 are 1.60, 1.56, and 2.03 percentage points with intervals
above zero, whereas the intervals at Top-1 to Top-5 include zero.
The gains over MSP-LLM are positive at every cutoff. Condition use thus
improves exact match mainly at deeper cutoffs, consistent with the
coverage effect reported in Section~\ref{exp: sop}.

\begin{table}[!htbp]
\centering
\caption{Paired differences in SOP exact match (percentage points) on the
918 test targets with reference precursors. Each cell reports the mean
difference and its 95\% bootstrap confidence interval; positive values
favor \proposed{}.}
\label{tab:app_sop_paired_ci}
\begingroup
\small
\setlength{\tabcolsep}{4pt}
\papertablesetup
\resizebox{\linewidth}{!}{%
\begin{tabular}{lrrrrrr}
\toprule
Comparison & Top-1 & Top-3 & Top-5 & Top-10 & Top-20 & Top-30 \\
\midrule
\proposed{} $-$ \proposed{} (w/o conditions)
& \makecell{+0.29\\{\scriptsize [$-$0.33, +0.91]}}
& \makecell{+0.98\\{\scriptsize [$-$0.04, +2.00]}}
& \makecell{+0.44\\{\scriptsize [$-$0.51, +1.42]}}
& \makecell{+1.60\\{\scriptsize [+0.29, +2.94]}}
& \makecell{+1.56\\{\scriptsize [+0.40, +2.76]}}
& \makecell{+2.03\\{\scriptsize [+0.65, +3.41]}} \\
\addlinespace[2pt]
\proposed{} $-$ MSP-LLM
& \makecell{+0.69\\{\scriptsize [+0.00, +1.42]}}
& \makecell{+1.20\\{\scriptsize [+0.07, +2.36]}}
& \makecell{+1.92\\{\scriptsize [+0.65, +3.27]}}
& \makecell{+3.96\\{\scriptsize [+2.29, +5.63]}}
& \makecell{+3.49\\{\scriptsize [+1.96, +5.05]}}
& \makecell{+2.54\\{\scriptsize [+0.98, +4.14]}} \\
\bottomrule
\end{tabular}}
\endgroup
\end{table}
\FloatBarrier
\subsection{Sensitivity to condition context at inference}
\label{app:condition_intervention}
Table~\ref{tab:app_condition_intervention} evaluates SOP generation alone
with reference precursors on all 918 test targets, keeping the
condition-trained checkpoints fixed and changing only the condition input
at inference (protocol in Appendix~\ref{app:evaluation_details}).
Removing the condition block or shuffling predicted conditions across
targets lowers exact match at both cutoffs; at Top-30, performance drops
from 47.89\% to 45.28\% and 44.73\%, respectively. Shuffled conditions fall
below the removed-block setting, indicating that the generator relies on
the content of the conditions and their alignment with the
target--precursor pair, not merely on their presence. This complements
Table~\ref{tab:main_condition_source}, which varies the condition source at
training and inference on the subset with recorded conditions.

\begin{table}[!htbp]
\centering
\caption{Sensitivity of SOP generation to condition context at inference
(exact match, \%; mean $\pm$ sample SD over three seeds).}
\label{tab:app_condition_intervention}
\begingroup
\small
\setlength{\tabcolsep}{6pt}
\papertablesetup
\begin{tabular}{lrr}
\toprule
Inference input & Top-10 & Top-30 \\
\midrule
Conditions removed & \msd{29.96}{0.33} & \msd{45.28}{0.45} \\
Conditions shuffled across targets & \msd{30.28}{0.95} & \msd{44.73}{1.09} \\
\midrule
Target-matched conditions & \msd{\textbf{32.03}}{0.66} & \msd{\textbf{47.89}}{0.38} \\
\bottomrule
\end{tabular}
\endgroup
\end{table}
\FloatBarrier

\FloatBarrier
\subsection{Condition context in candidate construction at 30 trials}
\label{app:planning_condition_control}
We compare \proposed{} with its variant without condition context in
candidate construction, using the same predicted precursor alternatives,
precursor masses, and utility-based selection procedure.
\proposed{} (w/o conditions) combines an SOP generator trained without
condition inputs with our precursor-overlap retriever without condition
weighting. Both retain the same rank-calibration procedure and fusion rule.
Conditions are removed only from SOP generation and retrieval;
condition predictions used for resource accounting remain unchanged.

\begin{table}[!htbp]
\centering
\caption{Candidate-construction control at 30 trials on the same 918
test targets. Recovery is reported as mean $\pm$ sample SD across
three SOP seeds. The final row reports the paired difference in
percentage points and its 95\% confidence interval.}
\label{tab:app_planning_condition_control}
\begingroup
\small
\setlength{\tabcolsep}{8pt}
\papertablesetup
\begin{tabular}{lr}
\toprule
Method & 30-trial recovery (\%) \\
\midrule
\proposed{} (w/o conditions) & \msd{37.87}{0.82} \\
\proposed{} & \msd{\textbf{40.01}}{0.35} \\
\midrule
Difference (pp) [95\% CI] & +2.14 [0.91, 3.41] \\
\bottomrule
\end{tabular}
\endgroup
\end{table}

At 30 trials, condition context improves recovery by 2.14 percentage
points (Table~\ref{tab:app_planning_condition_control}). We average
matched differences across the three SOP seeds within each target and
resample the 918 targets with replacement 20,000 times. The percentile
95\% interval is pointwise, without multiple-comparison adjustment, and
quantifies target sampling uncertainty conditional on the fitted models.
Differences are computed before rounding.
\FloatBarrier
\FloatBarrier
\subsection{Zero-shot synthesis planning}
\label{app:zero_shot_planning}
Table~\ref{tab:app_zero_shot} reports the GPT-5.4 mini zero-shot baselines
under the evaluation and resource-accounting protocols of
Tables~\ref{tab:main_msp_budgets} and~\ref{tab:main_unseen_trials}.
Values are recipe recovery (\%, mean $\pm$ sample SD over request seeds
42 and 46). Zero-shot recovery stays below 10\% on the original test set
and on single-reference new targets at every budget, and reaches 28.24\%
only on the 108 multi-reference new targets at 30 trials.

\begin{table}[!htbp]
\centering
\caption{Zero-shot synthesis planning with GPT-5.4 mini under trial and
furnace budgets. New (single) and New (multi) denote single-reference
($n=2{,}236$) and multi-reference ($n=108$) targets of the new set.}
\label{tab:app_zero_shot}
\begingroup
\small
\setlength{\tabcolsep}{4pt}
\papertablesetup
\resizebox{\linewidth}{!}{%
\begin{tabular}{llrrrrrrrrr}
\toprule
& & \multicolumn{3}{c}{Trials} & \multicolumn{3}{c}{Runs} & \multicolumn{3}{c}{Hours} \\
\cmidrule(lr){3-5}\cmidrule(lr){6-8}\cmidrule(lr){9-11}
Test set & Variant & 1 & 10 & 30 & 1 & 3 & 10 & 12 & 48 & 96 \\
\midrule
Original & end-to-end & \vsd{0.00}{0.00} & \vsd{0.93}{0.08} & \vsd{3.59}{0.31} & \vsd{3.27}{0.15} & \vsd{1.53}{0.15} & \vsd{3.49}{0.31} & \vsd{1.58}{0.08} & \vsd{2.72}{0.00} & \vsd{3.05}{0.15} \\
Original & PP$\rightarrow$SOP & \vsd{0.00}{0.00} & \vsd{2.23}{0.08} & \vsd{7.30}{0.31} & \vsd{6.54}{0.15} & \vsd{1.53}{0.31} & \vsd{7.24}{0.23} & \vsd{2.89}{0.39} & \vsd{4.58}{0.15} & \vsd{6.10}{0.15} \\
\addlinespace[2pt]
New (single) & end-to-end & \vsd{0.36}{0.13} & \vsd{1.63}{0.16} & \vsd{3.94}{0.06} & \vsd{2.42}{0.00} & \vsd{1.97}{0.06} & \vsd{3.49}{0.13} & \vsd{1.01}{0.03} & \vsd{2.57}{0.03} & \vsd{2.91}{0.06} \\
New (single) & PP$\rightarrow$SOP & \vsd{0.11}{0.03} & \vsd{2.59}{0.25} & \vsd{8.27}{0.13} & \vsd{5.75}{0.09} & \vsd{2.21}{0.28} & \vsd{8.25}{0.09} & \vsd{1.59}{0.16} & \vsd{4.85}{0.22} & \vsd{6.28}{0.47} \\
\addlinespace[2pt]
New (multi) & end-to-end & \vsd{1.39}{0.65} & \vsd{7.41}{0.00} & \vsd{14.81}{1.31} & \vsd{9.72}{1.96} & \vsd{10.65}{1.96} & \vsd{12.50}{0.65} & \vsd{4.63}{0.00} & \vsd{9.26}{1.31} & \vsd{10.19}{0.00} \\
New (multi) & PP$\rightarrow$SOP & \vsd{0.00}{0.00} & \vsd{15.28}{1.96} & \vsd{28.24}{5.89} & \vsd{18.52}{1.31} & \vsd{13.43}{3.27} & \vsd{27.78}{5.24} & \vsd{6.94}{0.65} & \vsd{17.13}{1.96} & \vsd{23.61}{4.58} \\
\bottomrule
\end{tabular}}
\endgroup
\end{table}
\FloatBarrier

\begingroup
\section{Limitations}
\label{app:limitations}

\paragraph{Evaluation scope.}
The available literature datasets do not provide experimental outcomes for
every proposed route. Recipe recovery therefore offers a reproducible
retrospective measure of whether a planner selects a recorded
precursor--operation pair within a specified budget. This directly tests
the prioritization of literature-supported routes, while leaving room for
valid alternatives absent from the references. A match does not establish
phase purity or synthesis success. The new target set broadens the chemistry and
reference coverage of this evaluation, although its references are
LLM-curated extractions without expert adjudication. Target disjointness
tests transfer to targets outside the task-specific training data; it does
not establish a temporal holdout or absence from foundation-model pretraining.

\paragraph{Resource-model scope.}
Public large-scale solid-state synthesis datasets
\citep{kononova2019text, lee2025text} encode materials, operations, and selected
reaction conditions, but do not provide the standardized process and
equipment metadata required for detailed laboratory scheduling.
In particular, ramp profiles, sample loads, crucible- and gas-specific
compatibility, and equipment overheads are not jointly annotated in their
released schemas. This data gap constrains the granularity at which
resource-aware synthesis planning can currently be benchmarked.
We therefore construct a common resource model from predicted operation
sequences and record-level predictions of temperature, duration, and
atmosphere. The formulation makes condition-dependent resource use an
explicit part of route selection and enables reproducible evaluation of
budget allocation when costs depend on shared furnace capacity.
All compared methods use the same compatibility and cost rules, enabling
controlled comparisons within these resource settings.
Repeating predicted maxima across stages, restricting sharing to the same
stage position, and using air/nonair classes simplify laboratory execution.
Capacity, overhead, and temperature tolerance are scenario assumptions;
predicted costs and compatibility are not guarantees of laboratory
feasibility. Resource settings and temperature-sharing sensitivity are specified
in Appendices~\ref{app:resource_planning_implementation}
and~\ref{app:temperature_milp}.
The formulation makes these assumptions explicit and provides a basis for
incorporating measured programs and equipment-specific constraints when
such data become available.

\paragraph{Candidate coverage and selection.}
Retrieval is limited to recorded sequences, and selection operates on the
upstream candidate pool. This finite-pool setting represents the decision
faced after candidate generation: how to allocate a limited budget among
the available alternatives. Candidate--selection cross-comparisons and
same-pool controls distinguish gains in candidate construction from gains
in allocation. Route utility expresses relative priority, so a certified
utility optimum need not maximize recipe recovery or synthesis success.
Planning is performed separately for each target and does not adapt to
experimental outcomes. Batch-processing platforms such as self-driving
laboratories \citep{szymanski2023autonomous} schedule experiments across
many targets and revise plans as results arrive; extending our formulation
to such settings requires target-level priorities for cross-target batches
and outcome-dependent updates for sequential campaigns. Both extensions
build on the same ingredients the current formulation already makes
explicit: predicted conditions as execution requirements, thermal stages
as the unit of furnace sharing, and route utility as the quantity to
allocate a budget against. The per-target allocation step established here
is therefore the component that a multi-target or closed-loop planner
would call at each round, rather than a separate problem.
\endgroup

\clearpage
\begingroup
\section{Prompt Templates}
\label{app:prompt_templates}
\newtcolorbox{promptbox}[1]{enhanced,breakable,
 colback=gray!3,colframe=blue!35!black,colbacktitle=blue!9,
 coltitle=black,coltext=black,fonttitle=\small\bfseries,
 title={#1},boxrule=0.5pt,arc=1.4mm,
 left=3mm,right=3mm,top=1.5mm,bottom=1.5mm,
 before skip=5pt,after skip=5pt}

\subsection{Joint condition prediction}
Both models use the following system message. Braced fields in the user
prompts are filled for each target--precursor pair.
\begin{tcolorbox}[colback=blue!3,colframe=blue!25,boxrule=0.4pt,arc=1mm]
\small\textbf{System}\par
You are an expert solid-state chemist planning a material synthesis.
\end{tcolorbox}

\begin{promptbox}{01\quad Joint condition prediction}
\textbf{User}\par\smallskip
{\ttfamily\footnotesize\raggedright [TASK: conditions] Given the target compound ['\{target\_formula\}'] and the selected precursors \{precursor\_list\}, predict the solid-state synthesis conditions: the maximum heating temperature (degrees Celsius), the maximum heating time (hours), and the heating atmosphere (one of ['air', 'nitrogen', 'oxygen', 'hydrogen', 'argon', 'other']).\par
Answer ONLY with a Python list of 'key=value' items in this order (include only the conditions you can predict): ['temp=<int>', 'time=<number>', 'atmos=<class>']. Example: ['temp=1200', 'time=12', 'atmos=air'].\par}
\end{promptbox}

\subsection{Condition-grounded operation prediction}
Temperature and time bounds are generated as described in
Appendix~\ref{app:retrieval_implementation}. The atmosphere clause is
omitted when the joint predictor does not emit atmosphere.
MSP-LLM uses the same template without the entire condition block.

\begin{promptbox}{02\quad Condition-grounded SOP}
\textbf{User}\par\smallskip
{\ttfamily\footnotesize\raggedright [CONDITIONS] temp in \{T\_low\}-\{T\_high\} C, time in \{t\_low\}-\{t\_high\} h, atmos=\{air\_or\_nonair\} [/CONDITIONS]\par
[PRECURSOR\_TYPES]\par
types\_present: \{precursor\_types\}\par
[/PRECURSOR\_TYPES]\par
\par
\par
Task:\par
Given the target compound ['\{target\_formula\}'], first classify its material group,\par
then based on the selected precursors \{precursor\_list\},\par
propose one plausible synthesis operation sequence using only the following operations:\par
['heating', 'sintering', 'annealing', 'mixing', 'shaping', 'quenching', 'drying'].\par
\par
Typically, two or more precursors are required.\par
\par
The possible material groups are:\par
['oxide', 'alloy', 'phosphate', 'pyrophosphate', 'halide', 'oxyhalide',\par
 'carbide', 'nitride', 'selenide', 'sulfide', 'other'].\par
\par
Output format:\par
[['material group'], ['precursor\_type1', 'precursor\_type2', ..., 'precursor\_typeK'], ['precursor1', 'precursor2', ..., 'precursorN'], ['op1', 'op2', ..., 'opM']]\par
\par
Return only the Python list of single-quoted strings with no extra text or explanation.\par}
\end{promptbox}
\endgroup

\end{document}